\documentclass[showpacs,aps,prd,superscriptaddres,notitlepage,showkeys,11pt]{revtex4-1}

\pdfoutput=1
\usepackage{times}
\usepackage{amssymb}
\usepackage{amsmath}
\usepackage{graphicx}
\usepackage{dcolumn}
\usepackage{hyperref}
\usepackage{xcolor}
\usepackage{mathptmx}
\usepackage{titlesec}
\usepackage{subcaption}

\begin{document}
\title{Mapping and Validating a Point Neuron Model on Intel's Neuromorphic Hardware Loihi}

\author{Srijanie Dey}
\affiliation{Department of Mathematics, Washington State University, Vancouver, WA, 98686}

\author{Alexander Dimitrov}
\affiliation{Department of Mathematics, Washington State University, Vancouver, WA, 98686}

\begin{abstract}
\textbf{Abstract} Neuromorphic hardware is based on emulating the natural biological structure of the brain. Since its computational model is similar to standard neural models, it could serve as a computational acceleration for research projects in the field of neuroscience and artificial intelligence, including biomedical applications. However, in order to exploit this  new generation of computer chips, rigorous simulation and consequent validation of brain-based experimental data is imperative. In this work, we investigate the potential of Intel's fifth generation neuromorphic chip - `Loihi', which is based on the novel idea of Spiking Neural Networks (SNNs) emulating the neurons in the brain. The work is implemented in context of simulating the Leaky Integrate and Fire (LIF) models based on the  mouse primary visual cortex matched to a rich data set of anatomical, physiological and behavioral constraints. Simulations on the classical hardware serve as the validation platform for the neuromorphic implementation.  We find that Loihi replicates classical simulations very efficiently and scales notably well in terms of both time and energy performance as the networks get larger.
\newline \textbf{Keywords -} Neuromorphic Computing, LIF, Neural Simulations, Validation, Performance Analysis
\end{abstract}

\maketitle

\section{Introduction}
The human brain is a rich complex organ made up of numerous neurons and synapses. Replicating the brain structure and functionality in classical hardware is an ongoing challenge given the complexity of the brain and limitations of hardware. The advent of supercomputers now allows for complex neural model, but at a huge cost of both software complexity and energy consumption.

A recent intense focus on brain studies, with the BRAIN  \cite{nihbrain} initiative at the US, the Human Brain Project (HBP) \cite{humanbrain} in Europe, and philanthropic endeavors like Janelia Research Campus \cite{janelia}, and the Allen Institute for Brain Science (AIBS) \cite{allenfirst}, has produced a wealth of new data and knowledge, from records of neuronal and network dynamics, to fine-grained data on network micro- and nano-structure, bringing in the era of big neural data. At the same time, advances in electronics and the search for post-von Neumann computational paradigms has led to the creation of neuromorphic systems like Intel’s Loihi \cite{loihi}, IBM’s TrueNorth \cite{truenorth1,truenorth2,truenorth3}, University of Manchester's SpiNNaker \cite{spinnaker} and HBP's BrainScaleS \cite{brainscale}.

Neuromorphic chips, as the name suggests - ‘like the brain’ - can mimic the brain's function in a truer sense as their design is analogous to the brain \cite{roy,questmimicbrain}. Inspired by its architecture, we work on developing a principled approach towards obtaining simulations of biologically realistic neural network models on a novel neuromorphic commercial hardware platform. 

\begin{figure}[htb]
\begin{center}
\includegraphics[width=0.5\linewidth]{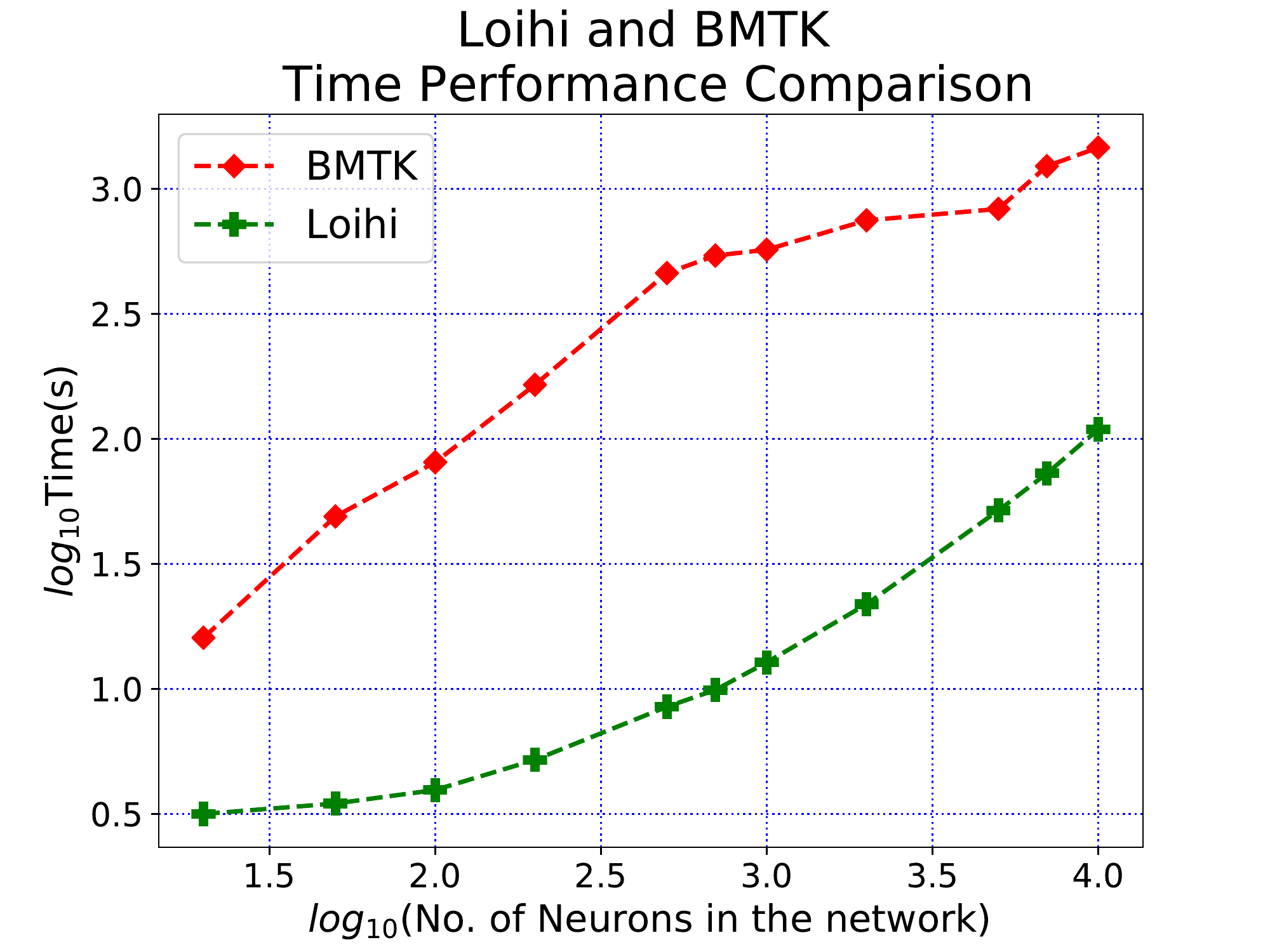}
\caption{As the network size increases, Loihi performs consistently both in terms of time and energy. Simulation of 500$\hspace{0.08cm}ms$ of dynamics for 10,000 neurons, results in a Loihi run-time of up to 109 seconds with maximum energy consumption of 5.7 Watts. (See Table \ref{table:1} for the explicit values.) }
\label{fig:time_energy}
\end{center}
\end{figure}

Computers today are limited in this respect because of the way they have been built historically and the way they process data leading towards more energy and resource consumption in order to maintain versatility \cite{von1,von2}. Neuromorphic chips on the other hand claim to be faster and more efficient for a set of specialized tasks \cite{Performance1,Performance2}.  
In this study, we try to validate this assertion based on neural models derived from the primary visual cortex (VISp) of the mouse brain, as seen in recent work done on SpiNNaker \cite{cortical,KnightSynapseCentric}.  Intel’s neuromorphic chip Loihi serves as our neuromorphic platform. Results obtained in Loihi are validated against classical simulations \cite{automaticfitting,allen_simu1,allen_simu2} given by AIBS's software package the Brain Modeling Toolkit (BMTK) \cite{bmtk}. Overall, our implementation indicates that Loihi is highly efficient in terms of time and energy in context of large brain network simulations and thus shapes our central motivation for this work (see Figure \ref{fig:time_energy} and Table  \ref{table:1}). The majority of this manuscript focuses on the trade-offs necessitated by these improvements, that is,  how precise are the Loihi simulations when validated against BMTK simulations?

\begin{center}
\begin{table}[htb]
\begin{tabular}{ |c | c | c  |} 
 \hline
 Network Size & Loihi Time(s) & BMTK Time(s) \\ 
 \hline
 20 & 3.16 & 16.08\\ 
 100 & 3.94 & 80.59 \\ 
 500 & 8.48 & 459.76\\
 1000 & 12.76 & 570.34\\
 5000 & 51.69 & 830.17 \\
 10000 & 109.02 & 1460.40\\
\hline
\end{tabular}
\caption{Time Usage in Loihi \& BMTK}
\label{table:1}
\end{table}
\end{center}

\vspace{-2cm}
As a first step in this approach, we focus on a class of neural network building blocks:  point neuronal models as used in large AIBS simulations. The Generalized Leaky Integrate and Fire Models (GLIFs) \cite{glif} have been found to be appropriate for reproducing cellular data under standardized physiological conditions. The data used for this study is made available by the AIBS.

The paper is organized as follows. In Section 2, we describe in detail the features of Loihi and the differences between the neuromorphic and classical hardware that form the basis for this study. Section 3 explains the implementation of the continuous LIF equation in BMTK on classical computational architecture, versus the discrete Loihi setting. Also, we list the validation methods and the cost function that is used to draw comparisons between the implementations.   In section 4, we list out and explain the various results leading to a qualitative and quantitative assessment between the two platforms based in part on methods from \cite{validation1}.

\section{COMPARISON BETWEEN CLASSICAL AND NEUROMORPHIC PLATFORMS}
At present, various simulators are available for implementing spiking neural networks \cite{review2008spikingneurons}. In this section, we lay out the details of the mathematical model and the platforms we use for our work. For the classical simulation, we use the Brain Modeling Toolkit (BMTK) \cite{bmtk} developed by the AIBS. Being open source, these resources enable us to experiment with a varied range of data and thus support our extensive validation of neuronal models in Loihi.  Intel's fifth-generation chip Loihi provides us with the tools to implement and test out the  various neuromorphic features. The output provided by Loihi simulations is then compared to the output of classical simulations implemented in BMTK.

\subsection{The Brain Modeling Toolkit (BMTK)}
The BMTK is a python-based software package for creating and simulating large-scale neural networks. It supports models of different resolutions, namely, Biophysical Models,  Point Models, Filter Models and Population Models along with the use of the rich data sets of the Allen Cells Database \cite{allenfirst}. It leverages the modeling file format SONATA which includes details on cell, connectivity and activity properties of a network along with being compatible with the neurophysiology data format Neurodata Without Borders (NWB), thus allowing easy access to a vast repertoire of experimental data.

In this study, we work with the Point Neuron Models with simulations supported by the BMTK module PointNet via NEST 2.11+ \cite{nest}. For analysis and visualization we use the HDF5 output format, underlying both SONATA and NWB's spike and time series storage.

\subsection{Loihi}
Hardware inspired by the structure and functionality of the brain, envisioned to provide advantages such as low power consumption, high fault tolerance and massive parallelism for the next generation of computers, is called neuromorphic hardware.
Towards the end of 2017, Intel Corporation unveiled its experimental neuromorphic chip called Loihi. We provide a summary of the platform here, since, unlike NEST and BMTK, Loihi is not so well-known in the computational neuroscience community. 

As of its 2020 rendition, Loihi is a 60-mm$^{2}$ chip that implements 131,072 leaky-integrate-and-fire neurons. According to \cite{loihi}, it uses an asynchronous spiking neural network (SNN), comprising of 128 neuromorphic cores, each with 1024 neural computational units; 3 x86 cores; along with several off-chip communication interfaces that provide connectivity to other chips. As Loihi advances the modeling of SNNs in silicon, it comprises of a large number of features necessary for their implementation viz., hierarchical connectivity, dendritic compartments, synaptic delays and synaptic learning rules. Each neuron is represented as a compartment in the Loihi architecture. The top-level microarchitecture of a Loihi neural computational unit resembles an actual biological neuron model comprising of all the functional units. The SYNAPSE unit processes all the incoming spikes from the previous compartment/neuron and captures the synaptic weight from the memory. The DENDRITE unit updates the different state variables. The AXON unit generates the spike message to be carried ahead by the fan out cores. The LEARNING unit updates the synaptic weights based on a learning rule.

The aim of this study is to establish the groundwork required to execute an ambitious plan of simulating about $\sim$250,000 neurons with $\sim$500M synapses, which encapsulates much of the experimentally observed dynamics in the mouse visual cortex available to the AIBS, thus providing a close functional replica of the mouse visual cortex. Loihi's specialized hardware features holds promise for a real-time, low-powered version of this implementation.

\subsection{Leaky Integrate and Fire Model (LIF)}
A typical neuron consists of a soma, dendrites and a single axon. Neurons send signals along an axon to a dendrite through junctions called synapses. The classical Leaky Integrate and Fire (LIF) equation \cite{book} is a point neuron model which reduces much of the neural geometry and dynamics in order to achieve computational efficiency. It is one of the simplest and rather efficient widely used representations of the dynamics of the neuron, and is given as:

\begin{align} 
    V'(t)&=\frac{1}{C}\left[I_{e}(t)-\frac{1}{R}(V(t)-E_{L})\right] \label{eq1}\\
    V(t)&\leftarrow{V_{r}} \text{, \hspace{0.1cm} if \hspace{0.1cm}} V(t)>\Theta \label{eq2}
\end{align}

where,
\begin{align*}
    V(t)&= \text{membrane potential (state)}\\
    C&= \text{membrane capacitance (parameter)}\\
    R&= \text{membrane resistance (parameter)}\\
    E_L&= \text{resting potential (parameter)}\\
    I_{e}&=\text{trans-membrane current (control and state)}\\
    V_{r}&=\text{reset membrane potential}\\
    \Theta&=\text{firing threshold}
\end{align*}

Here, $'=d/dt$, $t$ is time in $ms$, the membrane potential $V(t)$ of the neuron is in mV.  A LIF neuron fires when $V(t)>\Theta$, i.e. the membrane potential exceeds the firing threshold $\Theta$ and subsequently the membrane potential is set  to a reset value $V_{r}$.

 The classical LIF model (point generalized LIF) has been shown to
 match closely the dynamics of real neurons under a variety of conditions \cite{glif}, as listed in the Allen Cell Types Database \cite{allenfirst}. In addition, this model matches the LIF abstraction in Loihi to some extent. Thus, we work with this model throughout this study to establish the basis for comparison for the two platforms, validate the neuromorphic implementation against the ground truth of a standard implementation, and provide evidence that our neuromorphic platform performs more efficiently.

\subsection{Loihi LIF Model}

In an SNN, spiking neurons form the primary processing elements. The individual neurons are connected through junctions called synapses and interact with each other through single-bit events called spikes. Each spike train can be represented as a list of event times, e.g. as a sum of Dirac delta functions $\sigma(t)=\sum_{i}\delta(t-t_{i})$ where $t_{i}$ is the time of the $i$-th spike.

Since Loihi encapsulates the working of an SNN, one of the computational models it implements is a variation of the LIF model  based on two internal state variables : the synaptic current and the membrane potential \cite{loihi}. 
\begin{align}
    u(t)&=\sum_j w_j(\alpha_j*\sigma_j)(t)+b  \label{loihiU}\\
    \dot{v}(t)&=-\frac{1}{\tau_{v}}v(t)+u(t) \label{loihiV}\\
    v(t)&\leftarrow{0} \text{,\hspace{0.1cm} if \hspace{0.1cm} } v(t)>\theta \label{eq5}
\end{align}
where,
\begin{align*}
    v(t)&=\text{membrane potential}\\
    u(t)&=\text{synaptic current}\\
    w&=\text{synaptic weight}\\
    \alpha&=\text{synaptic response function}\\
    b&= \text{constant bias current}\\
    \theta&=\text{firing threshold}
\end{align*}

A neuron sends out a spike when its membrane potential exceeds its firing threshold $\theta$, i.e., $v(t)>\theta$. After a spike occurs, $v(t)$ is reset to $0$. As in the classical LIF model, here \hspace{0.12cm} $\dot{} =d/dt$. However, time and membrane potential values here are in arbitrary units.

Loihi follows a fixed-size discrete time-step model, similar to an explicit Euler integration scheme, where the time steps relate to the algorithmic time of the computation. This algorithmic time may differ from the hardware execution time. Moreover, to increase the efficiency of the chip, specific bit-size constraints are imposed on the state variables. We discuss the ones relevant for the LIF model implementation in the following section.

\section{METHODS}
\subsection{Model setup and integration}

The classical LIF model as represented in equation (\ref{eq1}) and (\ref{eq2}) can be rewritten as :

\begin{align}
    V'(t)&=-\frac{1}{\tau}V(t)+\frac{1}{C}\left[I_{e}(t)+\frac{1}{R}E_{L}\right]\label{eq6}
\end{align}

where $\tau=RC$ is membrane time constant of the neuron.

For a non-homogeneous linear differential equation,

\begin{align}
    \frac{df}{dt}=af+g \label{eq7}
\end{align}

the solution is given by the `variation of constants' method as :

\begin{align*}
    f(t)=e^{ct}\int_{0}^{t}g(s)e^{-cs}ds
\end{align*}

Comparing equation (\ref{eq6}) to equation (\ref{eq7}), we have,

\begin{align*}
    f &= V(t)\\
    g &= \frac{1}{C}(I_e)+\frac{1}{C}(E_L)
\end{align*}

 Here, the postsynaptic current $I_{e}$ is in the form of an exponent function. However, calculating the above integral at every step i.e. at all grid points $t_{i}\leq t$ proves to be quite expensive. 

BMTK uses NEST as backend to implement the above membrane potential dynamics. To avoid the expensive computations, NEST chooses to use the linear exact integration method \cite{nest_lif}, given below as follows :

Equation (\ref{eq6}) is rewritten as a multidimensional homogeneous differential equation:

\begin{align}
    \frac{d}{dt}y=Ay
\end{align}

where, \begin{equation*}
A = 
\begin{pmatrix}
a_{n} & a_{n-1} & \cdots & \cdots & a_{1} & 0  \\
1 & 0 & \cdots & 0 & 0 & 0 \\
0 & \ddots & \ddots & \vdots & \vdots & \vdots\\
\vdots  & \ddots  & \ddots & 0 & 0 & 0  \\
0 & 0 & \ddots & 1 & 0 & 0\\
0 & 0 & \cdots & 0 & \frac{1}{C} & \frac{1}{-\tau}
\end{pmatrix}
\end{equation*}

The solution is given by :

\begin{align} \label{eq:loihiInt}
    y(t)&=e^{At}y_{0}\\
    y_{t+h}&=y(t+h)=e^{A(t+h)}y(0)=e^{Ah}\cdot y_{t}
\end{align}

for a fixed time-step $h$. It saves exorbitant computations since each step involves multiplication only.

\subsubsection{Mapping between BMTK and Loihi models}

In this section, we illustrate the primary step of implementing the BMTK-NEST LIF integration into the fixed-size discrete time-step Loihi dynamics. Following the linear exact numerics in the NEST implementation, we implement our model in the Loihi discrete setting using the forward Euler method as discussed below.

First, we rewrite equation (\ref{eq1}), the standard LIF model to resemble the Loihi form as given in equation (\ref{loihiV}). Since Loihi parameters are unit-less, we introduce a re-scaling parameter $V_{s}$, which converts standard physical units used in BMTK to Loihi units.

As we compare equation (\ref{eq2}) and equation (\ref{eq5}), it can be seen that for BMTK the membrane potential reset value is set to $V_{r}$ whereas it is set to zero for Loihi. 

Thus, the forward transformation from BMTK to Loihi looks as follows :

\begin{align}\label{eq:forward}
    v=(V-V_{r})/V_{s}
\end{align}

which produces an inverse transformation, to arrive back at the BMTK values, given by :

\begin{align}\label{eq:inverse}
    V=v\cdot V_{s}+V_{r}
\end{align}

Substituting the expression in (\ref{eq:inverse}) in (\ref{eq1}) and isolating $v$, we get :

\begin{align} 
    V'(t)&=\frac{1}{C}\left[I_{e}(t)-\frac{1}{R}(V(t)-E_{L})\right] &| V = v V_s + V_r \implies\\
    v'(t) V_s &= -\frac{1}{R C} (v V_s + V_r - E_L) + \frac{1}{C} I_e(t) &| /V_s \implies \\
    v'(t) &= -\frac{1}{\tau_v} v(t) + \frac{1}{\tau_v}\frac{E_L-V_r}{V_s} + \frac{1}{C}\frac{I_e}{V_s} \\
    & = -\frac{1}{\tau_v} v(t) +u(t)
\end{align}
with 

\begin{align}
    v(t) &= \frac{V(t) - V_r}{V_s},\\
    u(t) &= \frac{1}{C V_s} I_e(t) + \frac{1}{\tau_{v}} \frac{E_L - V_r}{V_s},\\
    \tau_v &= R C, \\
    \theta &= \frac{\Theta - V_r}{V_s} \label{eq:thres}
\end{align}
Here, we reintroduce the LIF threshold $\Theta$ and the corresponding  Loihi threshold $\theta$ in equation (\ref{eq:thres}).

Loihi implements the continuous LIF as a discrete finite state machine model \cite{izhi,fixedpointspinnaker} implemented in silicon. The actual computation is similar to a forward Euler scheme with some peculiarities reflecting engineering design trade-offs. Specifically, the $v(t)$ state evolves on-chip according to the update rule,

\begin{align}
    v(t+1)=v(t)\left[1-\frac{\delta_{v}}{2^{12}}\right]+b+u(t)\label{eq21}
\end{align}

where $\delta_{v}$ is the membrane decay constant.

Using the forward Euler method :

\begin{align*}
    y_{n+1}=y_{n}+f(t_{n},y_{n}).dt
\end{align*}

where $y_{n+1}=y(t_{n+1})$ and $t_{n+1}=t_{n}+dt$ for a fixed time-step $dt$, we transform the classical LIF model into a form followed in equation (\ref{eq21}). Thus, transforming the LIF model into the discrete form and grouping terms to match the Loihi integration (\ref{eq:loihiInt}) yields the following : 

\begin{align}
    \frac{v(t+dt)-v(t)}{dt} &=-\frac{1}{\tau_{v}}v(t)+u(t)  \\
   \implies v(t+dt) &= v(t)(1-\frac{dt}{\tau_v}) + u(t) dt\label{loihiVdt}
\end{align}

where $dt$ is the fixed time-step. 

In order to equate the Loihi computation (\ref{eq21}) with the Euler scheme (\ref{loihiVdt}),  we use $dt$ with units \emph{ms/Loihi timestep} i.e. 1 BMTK millisecond per Loihi timestep that leads to a Loihi voltage decay parameter $\delta_v$ such that,
\begin{align}\label{eq:tau}
    (2^{12}-\delta_v)~ 2^{-12} &= (1-\frac{dt}{\tau_v}) \\
    \implies \delta_v &= \frac{dt}{\tau_v}~2^{12} = \frac{dt}{R C}~2^{12}
\end{align}

\subsection{Validation Methods}

\begin{itemize}

\item[]{\textbf{Data}}\\
The data used here is provided by the Allen Mouse Brain Atlas \cite{allenfirst}, which is a survey of single cells from the mouse brain, obtained via intracellular electrophysiological recordings done through a highly standardized process. We focus on neurons with different morphologies with available GLIF parameters. The data used can be accessed in the Allen Cell Types Database. Our LIF model is implemented and simulated on BMTK based on this data, and these simulations form the ground truth for validating the Loihi implementations.

\item[]{\textbf{Methods and Cost Function}} \\
A spike-to-spike equivalence between the simulated spiking activity and the experimental data only allows for a qualitative validation. Since BMTK and Loihi run on two different computing environments, a statistical comparison becomes imperative to quantify the level of similarity between the two implementations.  

We compare the simulation dynamics for both implementations based on the following - 

\begin{enumerate}
    \item[] \textbf{Raster Plot:}  We evaluate the membrane potential response at each time-step. The X-axis represents the membrane potential and the Y-axis represents the time-step. Raster plot helps to highlight the difference in the spike dynamics at each step and thus tune the parameters as needed.
    \item[] \textbf{Scatter Plot:}  For examining association between the two implementations, we use color-coded scatter plots identifying the correlation relationships. We add a trend line to illustrate the strength of the relationship and pin down the outliers to improve the simulation results. Since we anticipate an almost perfect linear relationship, we quantify the match with its Pearson correlation coefficient,
    \begin{align}
        r = \frac{\sum_{i=1}^n \left({y}_{L}^{i}-\bar{y_{L}}\right)\left({y}_{B}^{i}-\bar{y_{B}}\right)}{\sqrt{\sum_{i=1}^n \left({y}_{L}^{i}-\bar{y_{L}}\right)^2\sum_{i=1}^n\left({y}_{B}^{i}-\bar{y_{B}}\right)^2}}
    \end{align}
    where,
    \begin{align*}
        i&=\text{index of data point}\\
        {y}_{L}&=\text{transformed Loihi values}\\
        y_{B}&=\text{original BMTK values}\\
        n&=\text{number of data points}
    \end{align*}
    \item[] \textbf{Distribution  Function:}  Another approach we take is to compare the distributions of attained state values in the two cases. We use density plot as a representation of those distributions, thus allowing us to compare the two implementations in terms of concentration and spread of the values and provide a basis for comparing the collective dynamics of the implementations. 
    \item[] \textbf{Cost Function:} To quantify the error between the BMTK and Loihi membrane potential values, we base our cost function on Root Mean Square Error (RMSE) values as follows :
    \begin{align}
    RMSE = \sqrt{\frac{1}{n}\sum_{i=1}^{n}\left({y}_{L}^{i}-y_{B}^{i}\right)^{2}}
    \end{align}
\end{enumerate}
\end{itemize}

\section{RESULTS}
In order to be able to simulate a network of over 250,000 neurons with a connectivity of over 500M synapses in the neuromorphic hardware, we begin by ensuring an high quality replication of individual neural and smaller network models. The replication performance here is evaluated based on membrane potential and current responses, the two state variables. We conjecture that securing a good replica for smaller models will ensure that parameters can be calibrated correctly and thus can be carried forward for the bigger networks needed in biological context \cite{validation1, validation2, singleneuronreview}. We begin our work on a single-neuron network sub threshold dynamics driven by both bias current and external spikes to ensure Loihi is able to handle both stimuli efficiently. Our test suite consists of LIF models based on 50 different parameter sets. We perform rigorous quantitative analysis of our results based on various statistical measures to demonstrate that we have replication of high quality. It is important to restate here that we test our results based on neurons with different morphologies and biophysics, which attribute to the different parameter sets.

\subsection{Simulations of a Single Neuron}
We begin by simulating a single-neuron network in BMTK. The simulation is run for 500$\hspace{0.05cm}ms$. The classical parameters are translated to Loihi values and the corresponding LIF model is implemented as an one-neuron SNN executed for 500 time steps in Loihi.  The simulations are driven either by bias current or external spikes. In the Loihi network, neurons are denoted by compartments. The compartment dynamics are guided by the parameters - bias current mantissa, membrane potential threshold, membrane potential decay and current decay. It is worth iterating here that the membrane potential values in Loihi are unit-less as opposed to the BMTK values which are assigned units of millivolts (mV).
We test the efficacy of our replication, both qualitatively and quantitatively, for all 50 parameters sets and find that the results are consistent with the ones described below. 

\begin{figure}[h]
\begin{center}
\includegraphics[height=1.7in,width=4.0in]{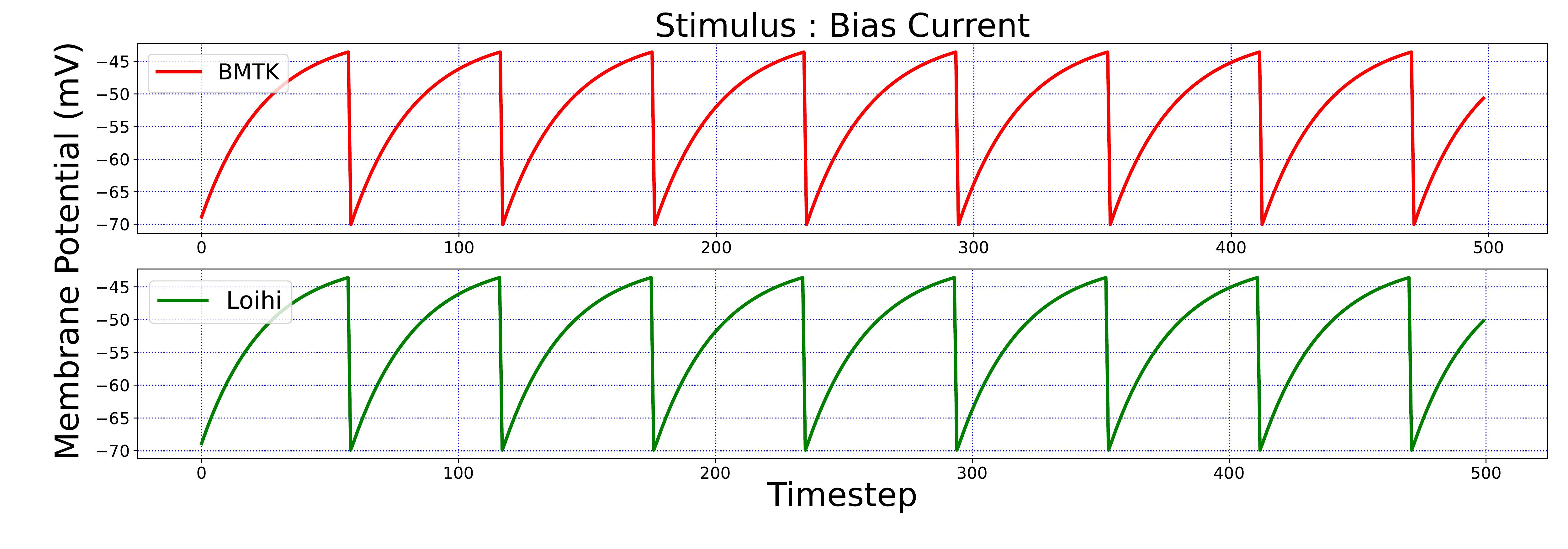}\\
\includegraphics[height=1.7in,width=4.1in]{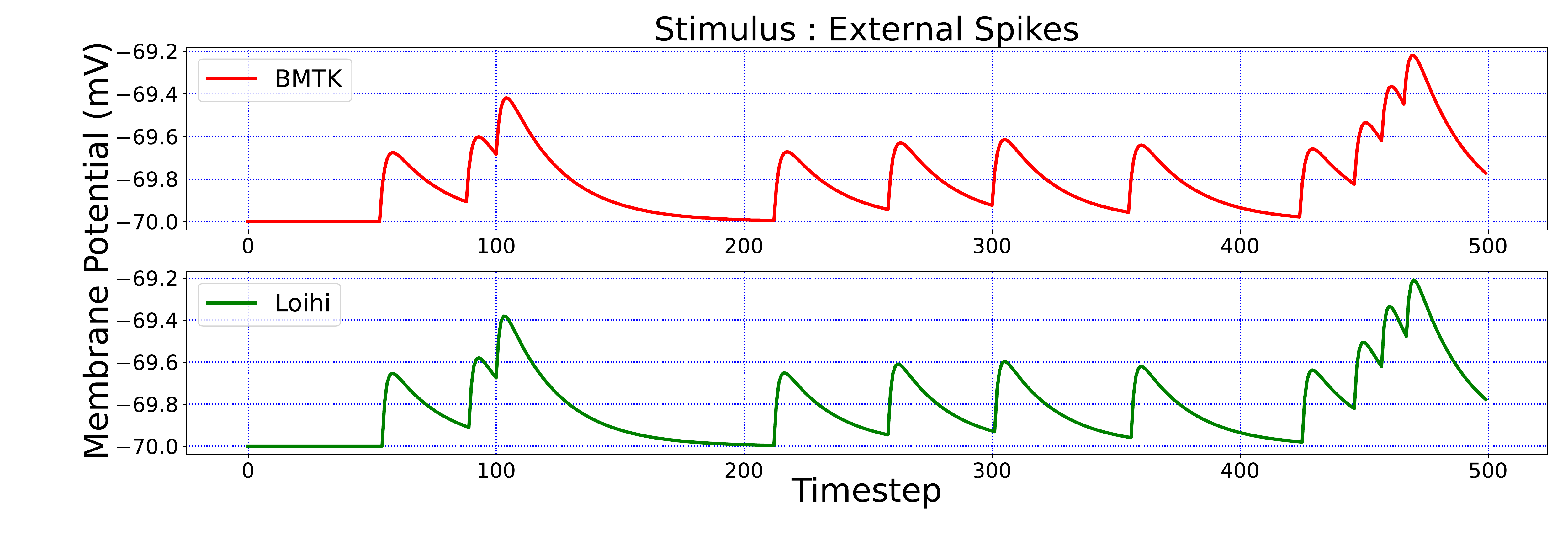}\\
\caption{Membrane potential response for single-neuron network based on two different neuron parameters. The first simulation is driven by bias current and the second one by external spikes.}
\label{fig:Single_Neuron}
\end{center}
\end{figure}

In Figure \ref{fig:Single_Neuron}, we illustrate the implementations achieved through bias current and external spikes on two different parameter sets.

For a qualitative comparison, it can be seen from Figure \ref{fig:Single_Neuron} that Loihi implementations simulate BMTK results very closely. We have exactness in terms of spike frequency, spike amplitude, and response values. Since Loihi membrane potential values are unit-less, we map them back to the BMTK values (mV) before performing the comparison. The inverse mapping from Loihi to BMTK is performed based on equation (\ref{eq:inverse}), i.e.,
\begin{equation*}
     V=v\cdot V_{s}+V_{r}
\end{equation*}

We perform a quantitative assessment of the replication using  several statistical measures. As seen from Table \ref{table:2}, the values on the two platforms are highly correlated with a relatively small RMSE.

\begin{center}
\begin{table}[h!]
\begin{tabular}{|c |c| c|} 
 \hline
 Stimulus & Correlation & RMSE  \\ 
 \hline
 Bias Current & 0.999992 & $1.1374 \times 10^{-4}$  mV/$ms$  \\ 
 External Spikes & 0.999942 & $4.208 \times 10^{-5}$ mV/$ms$\\ 
 \hline
\end{tabular}
\caption{Correlation and RMSE between BMTK and Loihi membrane potential values}
\label{table:2}
\end{table}
\end{center}

\begin{figure}[htb]
    \begin{minipage}[t]{.3\textwidth}
        \centering
        \includegraphics[width=\textwidth]{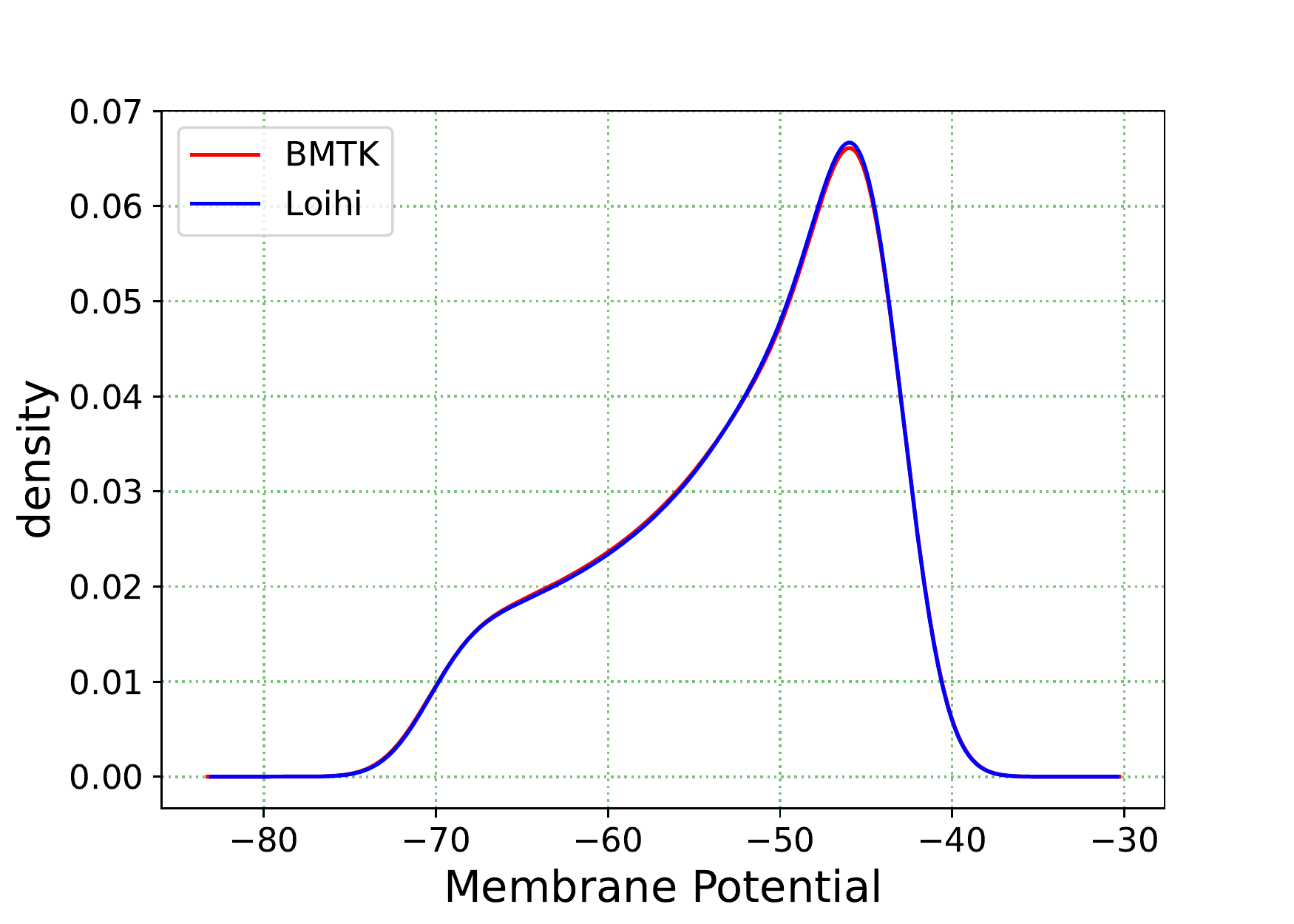}
    \end{minipage}
    \hfill
    \begin{minipage}[t]{.3\textwidth}
        \centering
        \includegraphics[width=\textwidth]{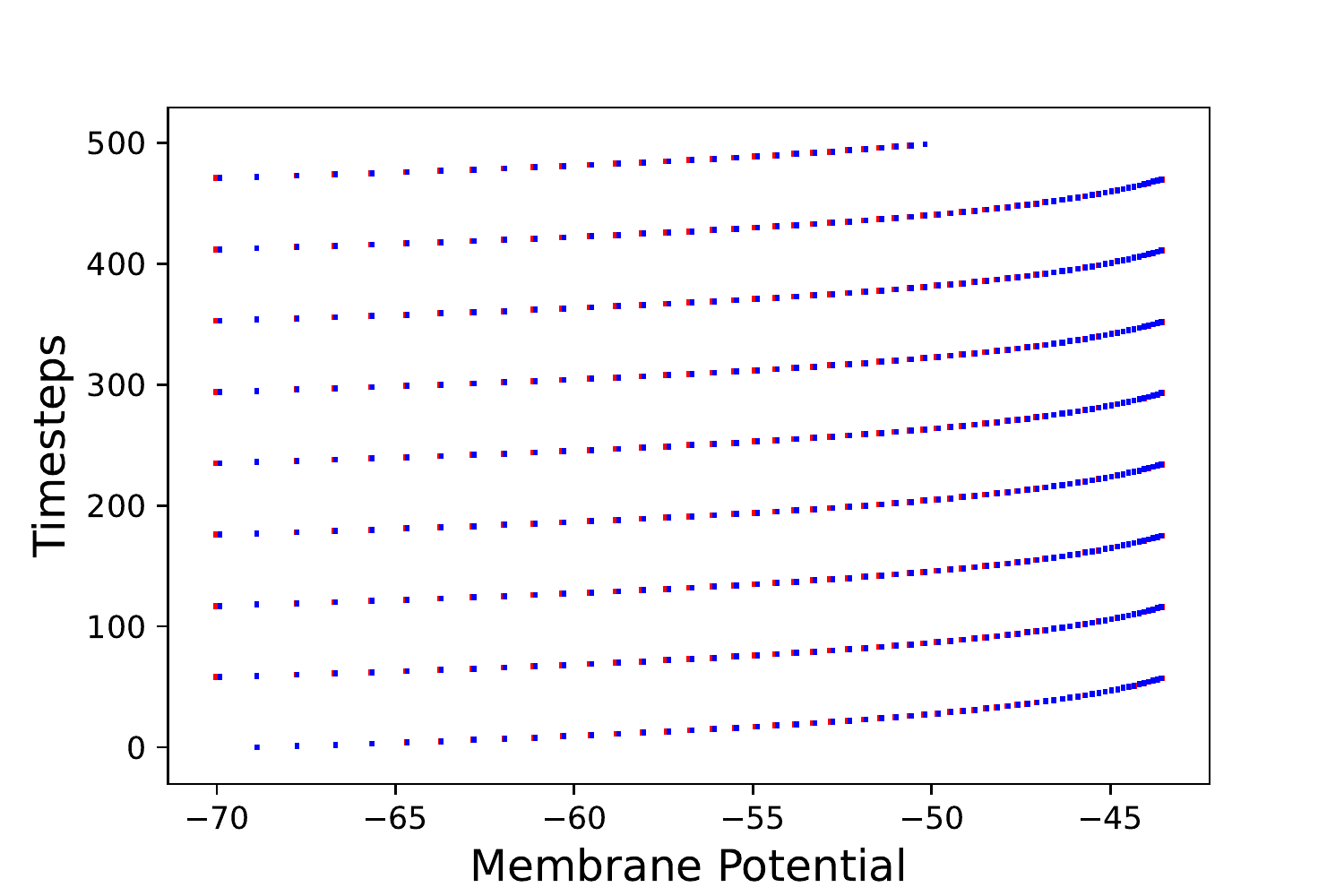}
    \end{minipage}
    \hfill
    \begin{minipage}[t]{.3\textwidth}
        \centering
        \includegraphics[width=\textwidth]{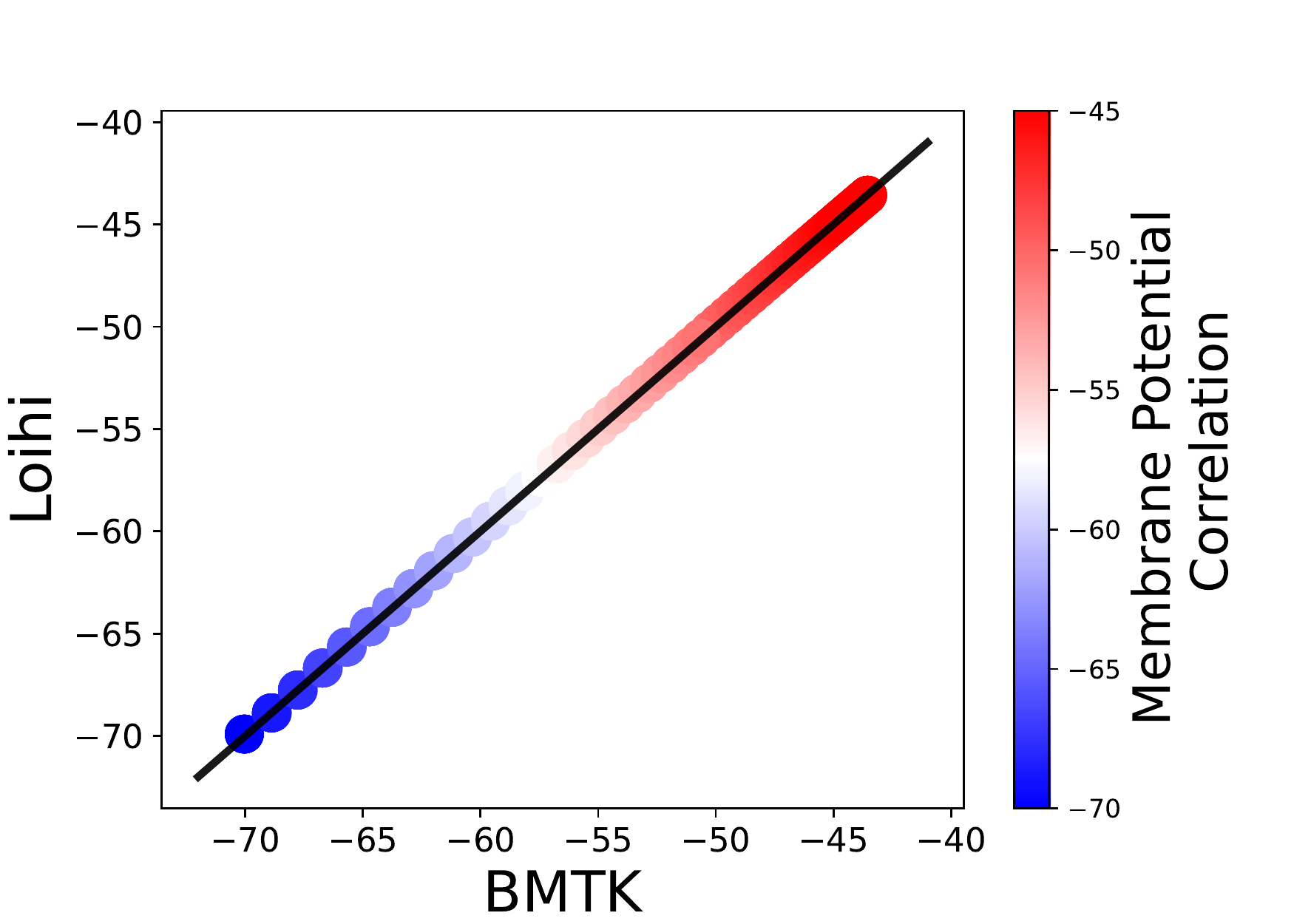}
    \end{minipage}  
    \newline
    \begin{minipage}[t]{.3\textwidth}
        \centering
        \includegraphics[width=\textwidth]{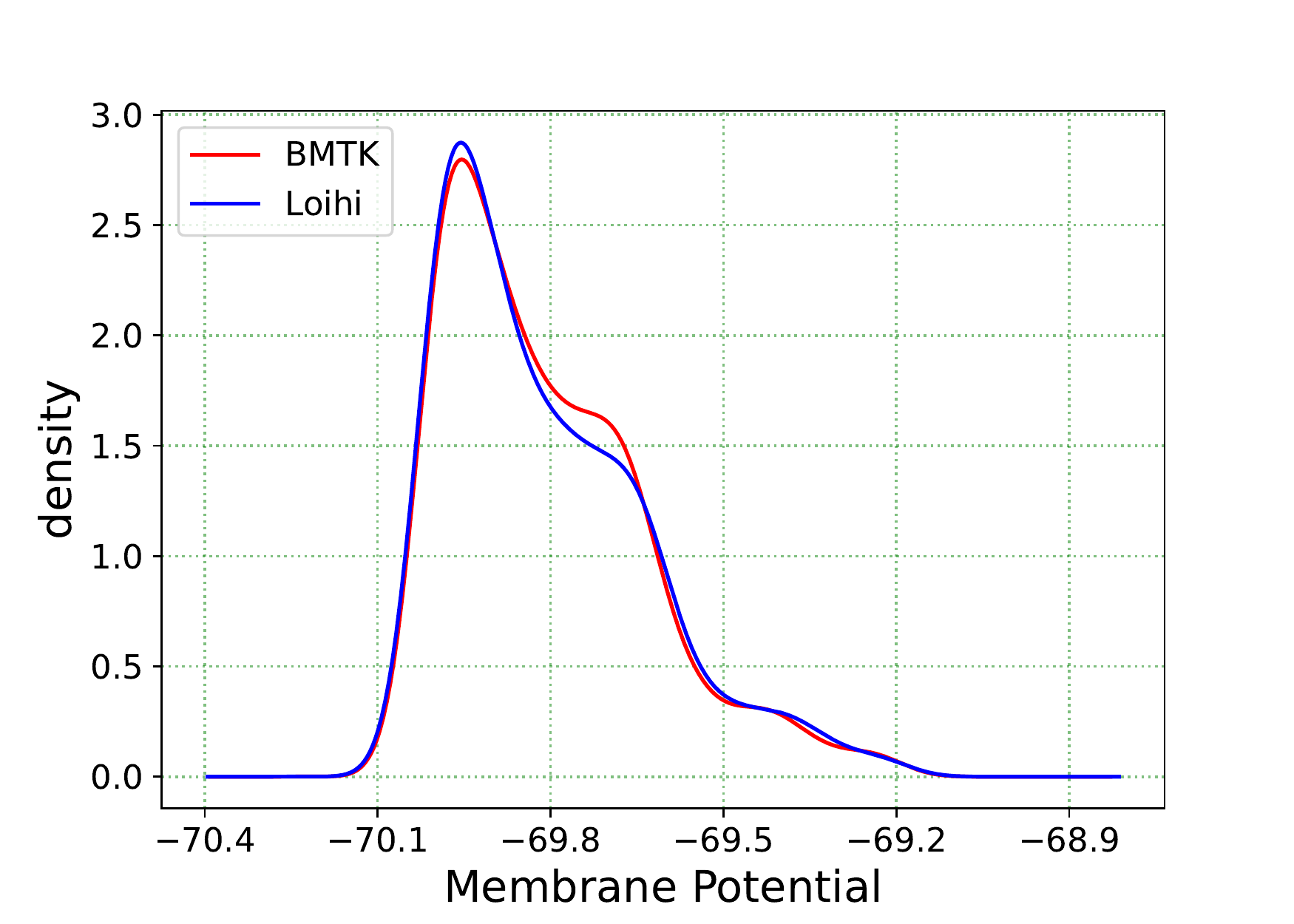}
        \subcaption{Density Plot}
    \end{minipage}
    \hfill
    \begin{minipage}[t]{.3\textwidth}
        \centering
        \includegraphics[width=\textwidth]{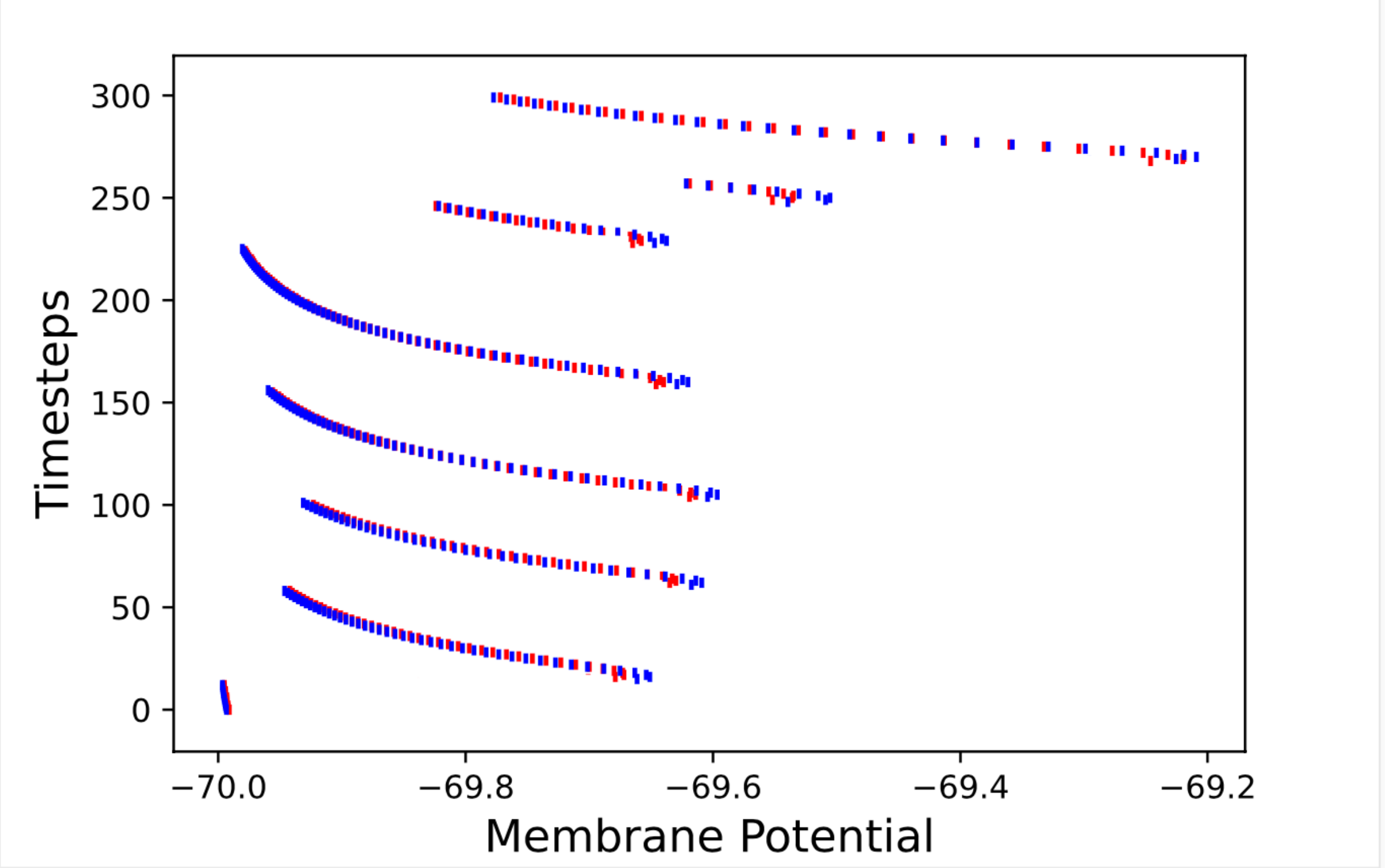}
        \subcaption{Raster Plot}
    \end{minipage}
    \hfill
    \begin{minipage}[t]{.3\textwidth}
        \centering
        \includegraphics[width=\textwidth]{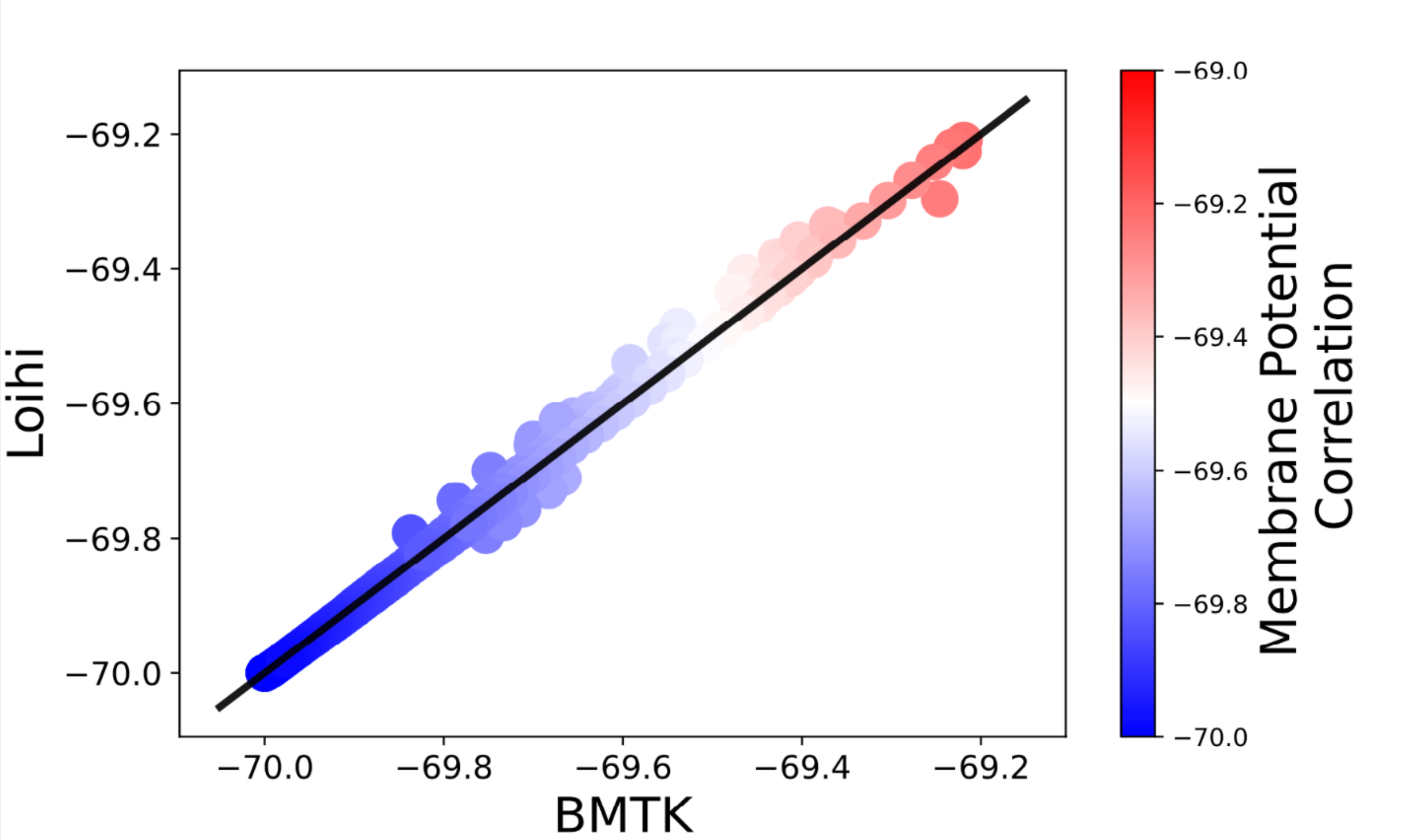}
        \subcaption{Scatter Plot}
    \end{minipage}  
    \caption{Validation Plots for simulations based on two different stimuli. The first and the second row show the results for Bias Current and External Spike stimulus respectively.}
    \label{fig:VM}
\end{figure}

Figure \ref{fig:VM} illustrates the  comparison of Loihi implementations against the BMTK implementations for the two different stimuli using various statistical measures - (a) Distribution function approximating the membrane potential dynamics (b) Raster plot of the spiking network activity  (c) Scatter Plot highlighting the positive coefficient between the two implementations.

\subsection{Simulation Using Varied Precision}
As already stated, Loihi follows a fixed-size discrete time-step model along with bit-size constraints for the different parameters. Thus, we investigate how changing the precision of the time scale and the neuron state values affects the accuracy of the simulations. We investigate this property for the two state variables - membrane potential and current. 

Figure \ref{fig:BMTK} below illustrates the membrane potential and current responses of a single neuron model in BMTK which form the basis of our comparison for the results below. 

\begin{figure}[htb]
   \begin{minipage}[t]{\textwidth}
        \centering
        \includegraphics[width=0.7\textwidth]{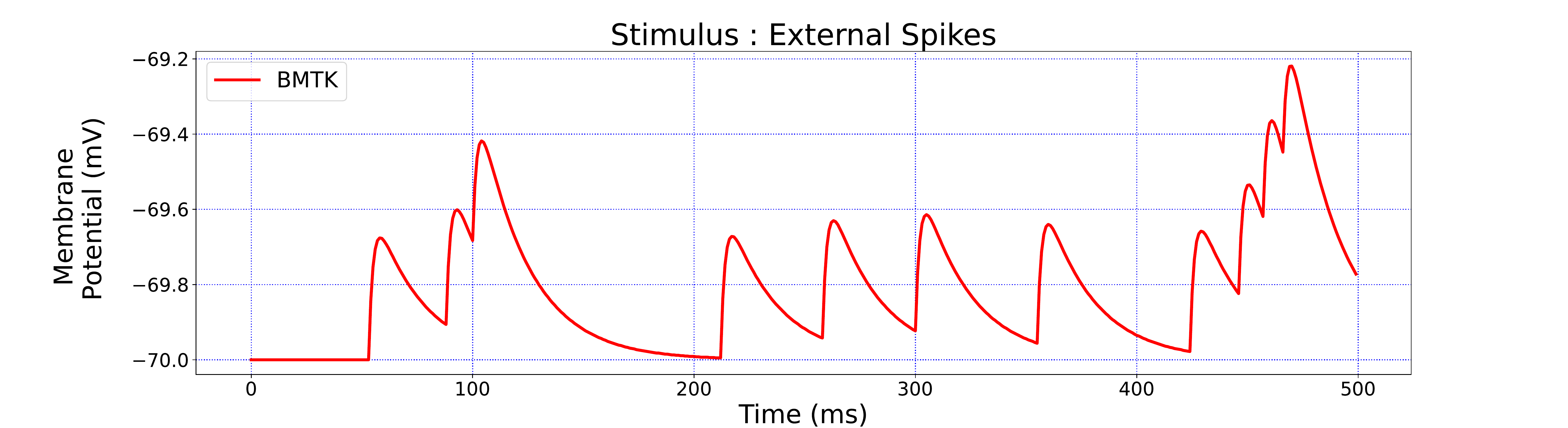}
    \end{minipage}  
    \newline
    \begin{minipage}[t]{\textwidth}
        \centering
        \includegraphics[width=0.7\textwidth]{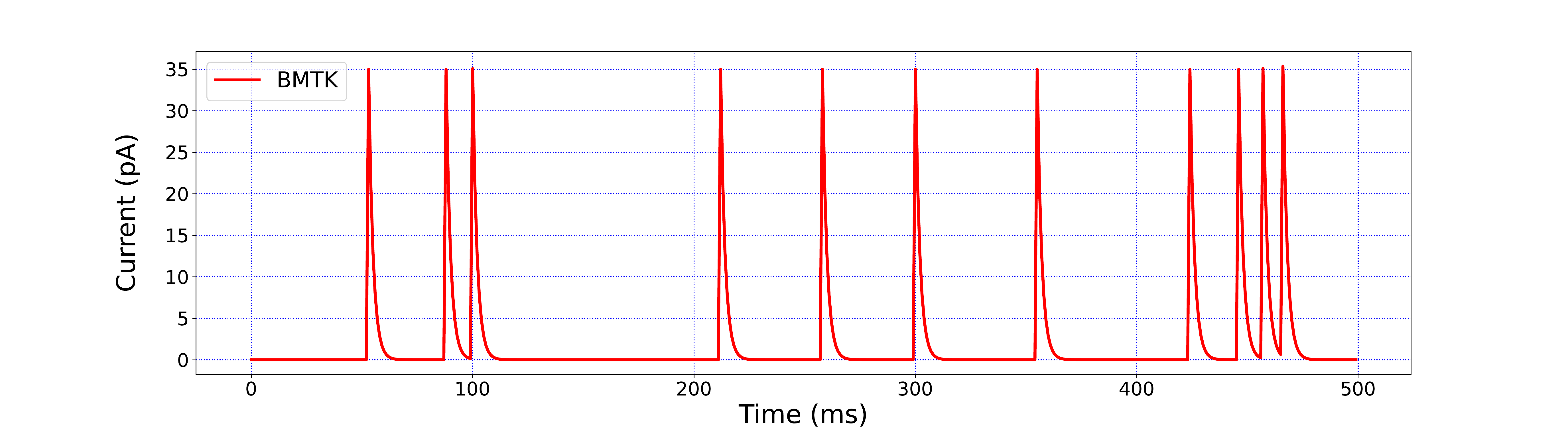}
    \end{minipage}
    \caption{Membrane potential response and  current response for a single neuron model in BMTK}
    \label{fig:BMTK}
\end{figure}

\subsubsection{Simulation Using Varied Temporal Precision}
For Loihi's fixed simulation time-step, we assign different time units to each step and test the corresponding simulation precision. This is achieved through the `$dt$' available in our equations while transforming the classical LIF model to Loihi neural model. It enables us to experiment with several time units \cite{Hopkins2015}. Following equation (\ref{eq:tau}), the change of a time-step while working with the Loihi neural  model necessitates a corresponding variation of the time constant `$\tau$' to yield the desired results.

\begin{figure}[htb]
   \begin{minipage}[t]{0.48\textwidth}
        \centering
        \includegraphics[width=\textwidth]{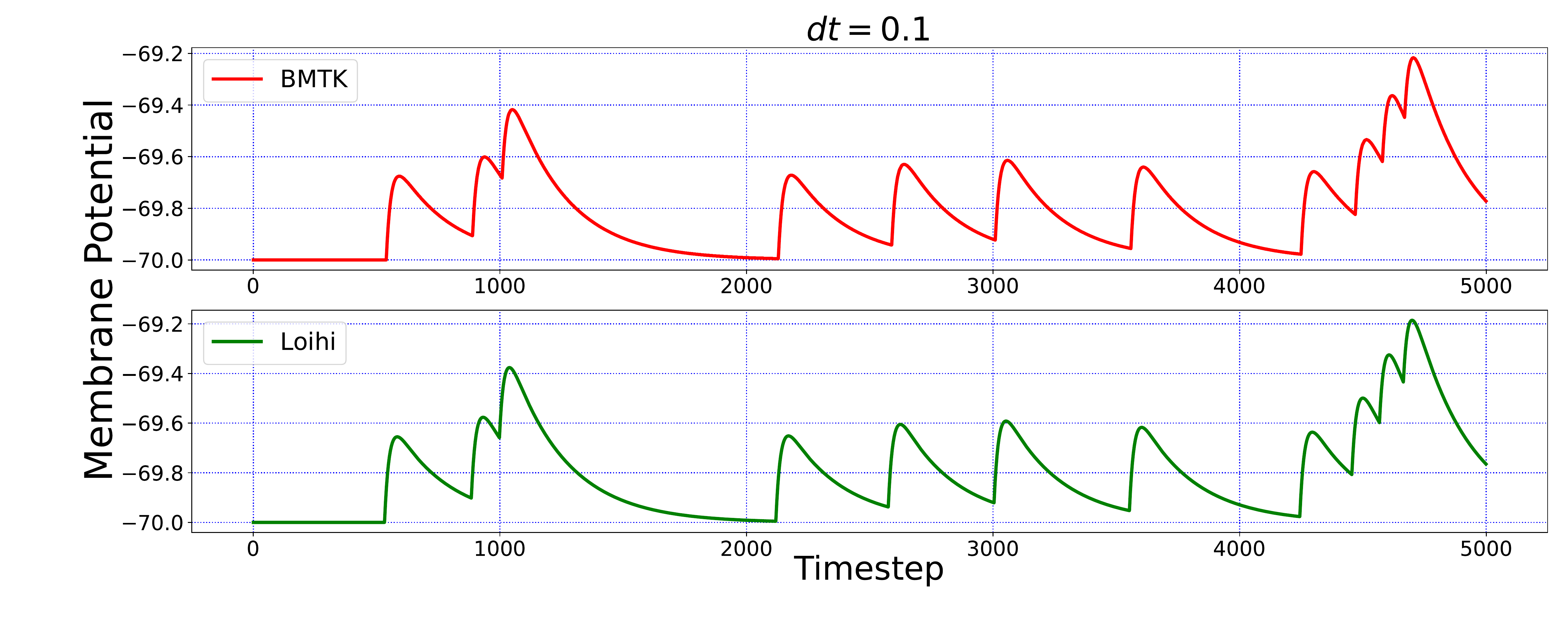}
    \end{minipage}  
    \hfill
    \begin{minipage}[t]{0.48\textwidth}
        \centering
        \includegraphics[width=\textwidth]{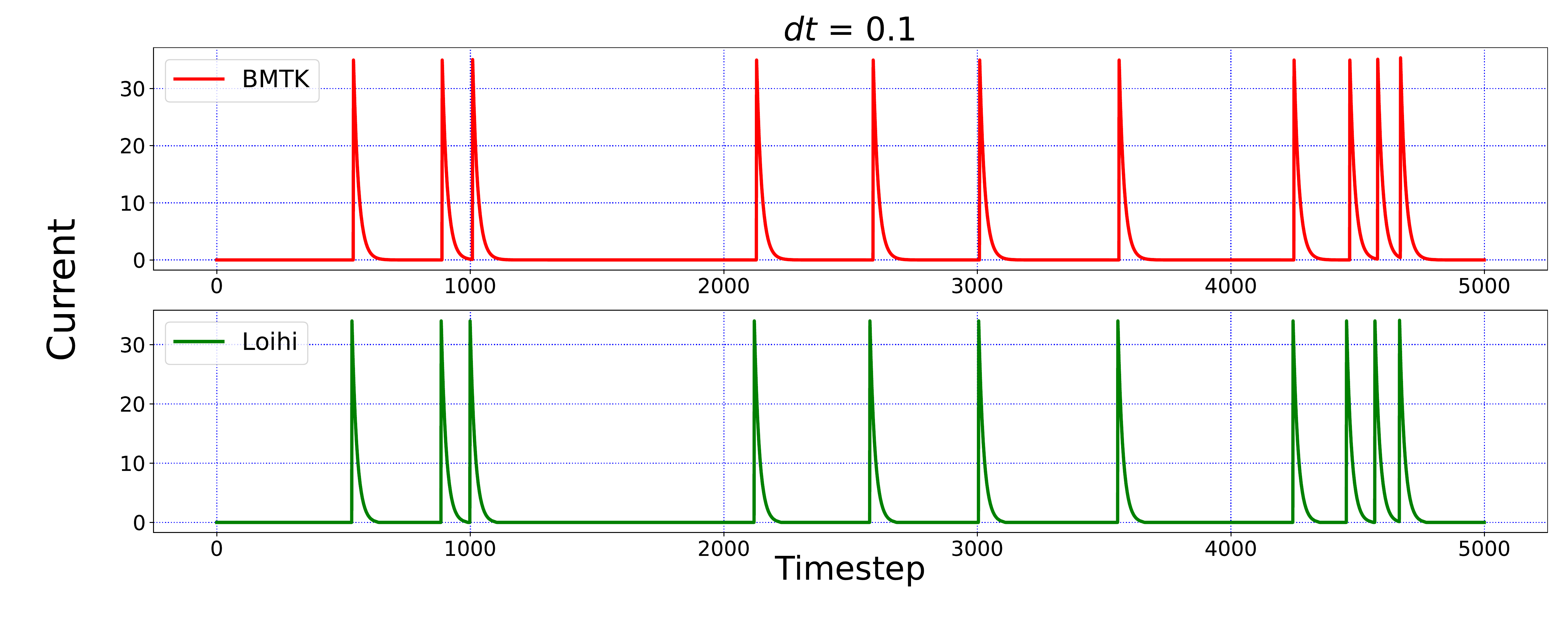}
    \end{minipage}
    \newline
    \begin{minipage}[t]{0.48\textwidth}
        \centering
        \includegraphics[width=\textwidth]{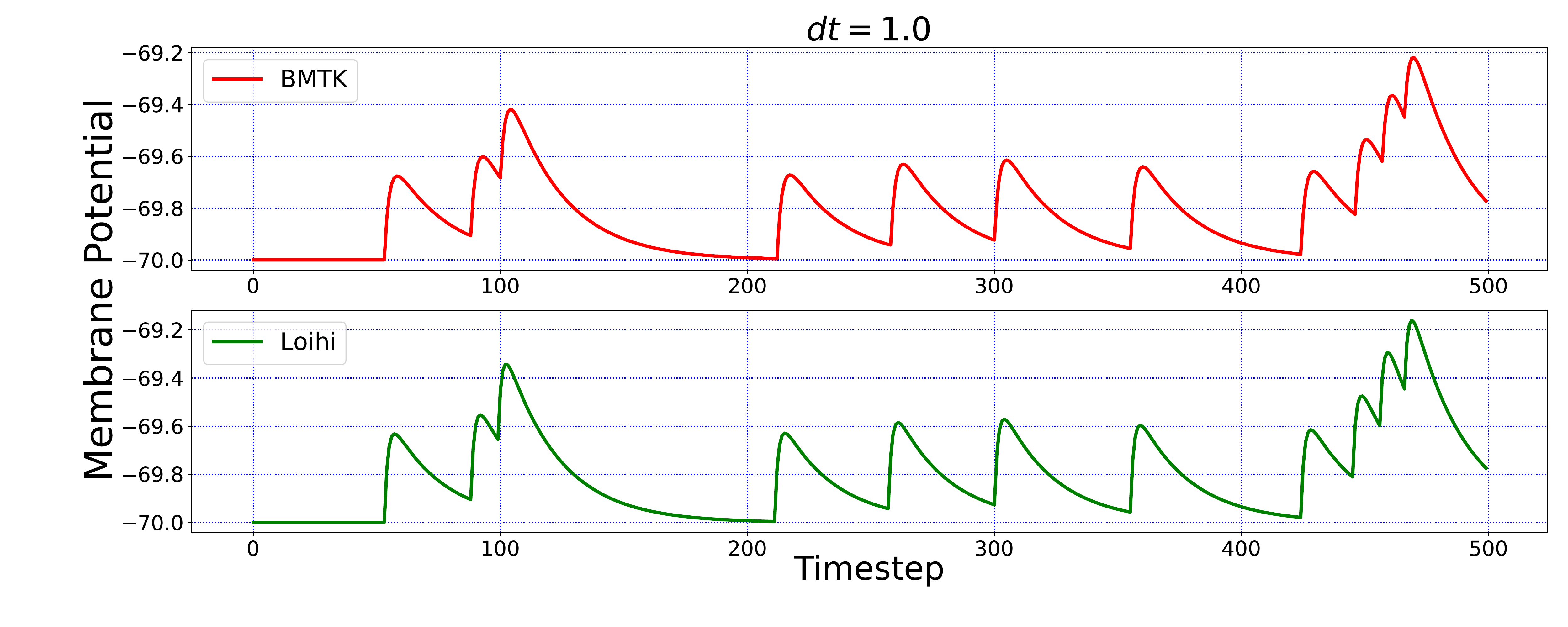}
    \end{minipage}  
    \hfill
    \begin{minipage}[t]{0.48\textwidth}
        \centering
        \includegraphics[width=\textwidth]{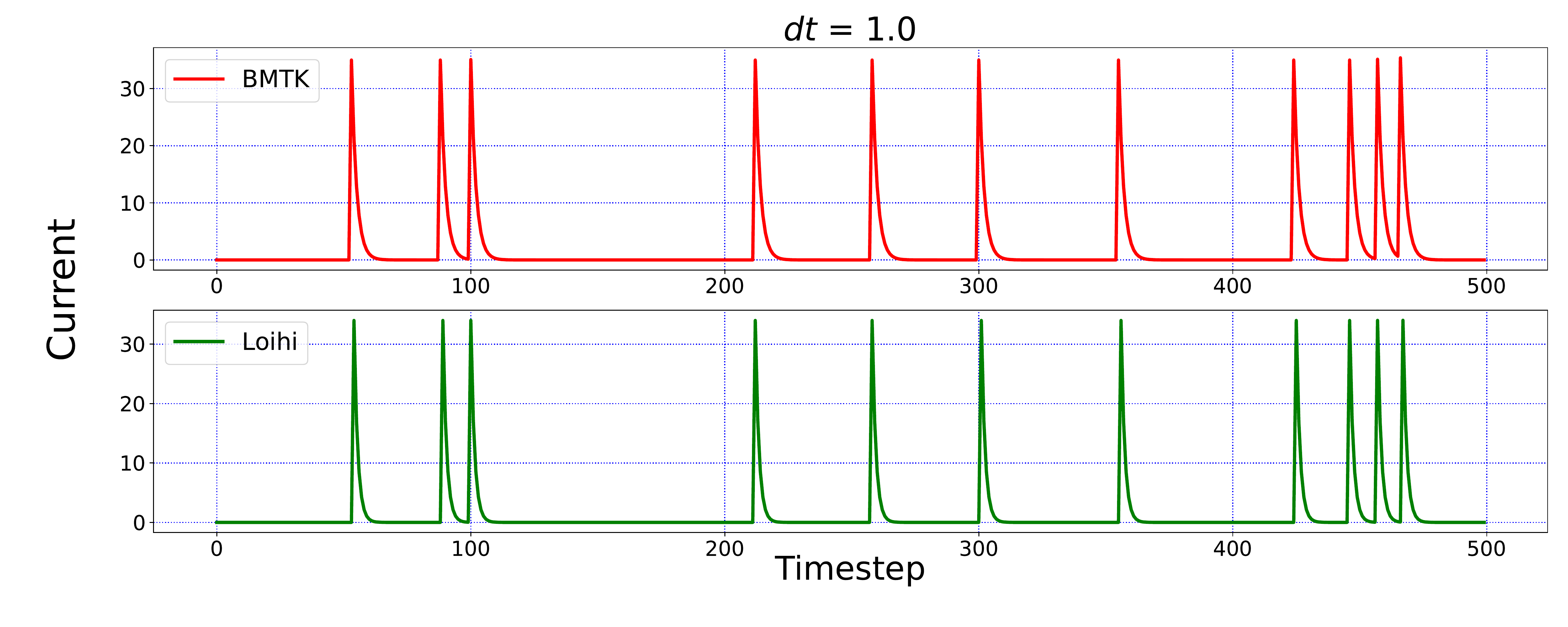}
    \end{minipage}
    \newline
    \begin{minipage}[t]{0.48\textwidth}
        \centering
        \includegraphics[width=\textwidth]{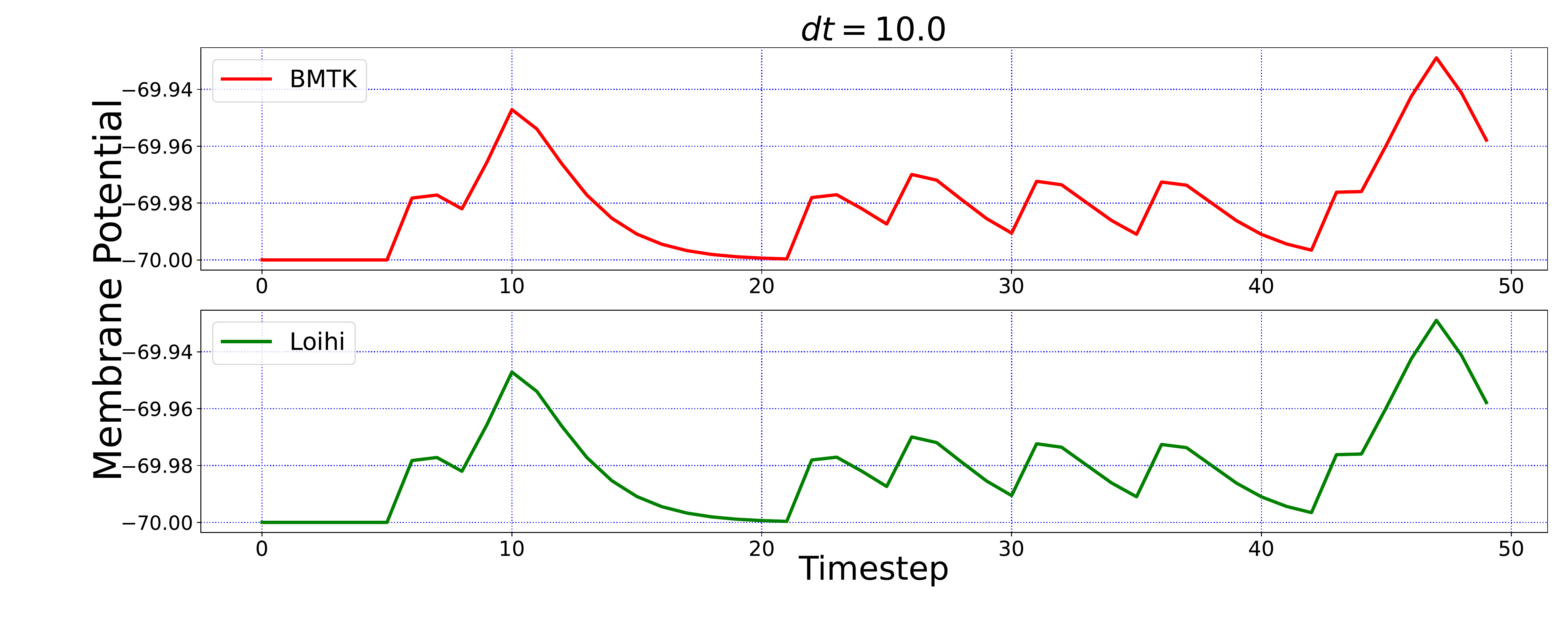}
    \end{minipage}  
    \hfill
    \begin{minipage}[t]{0.48\textwidth}
        \centering
        \includegraphics[width=\textwidth]{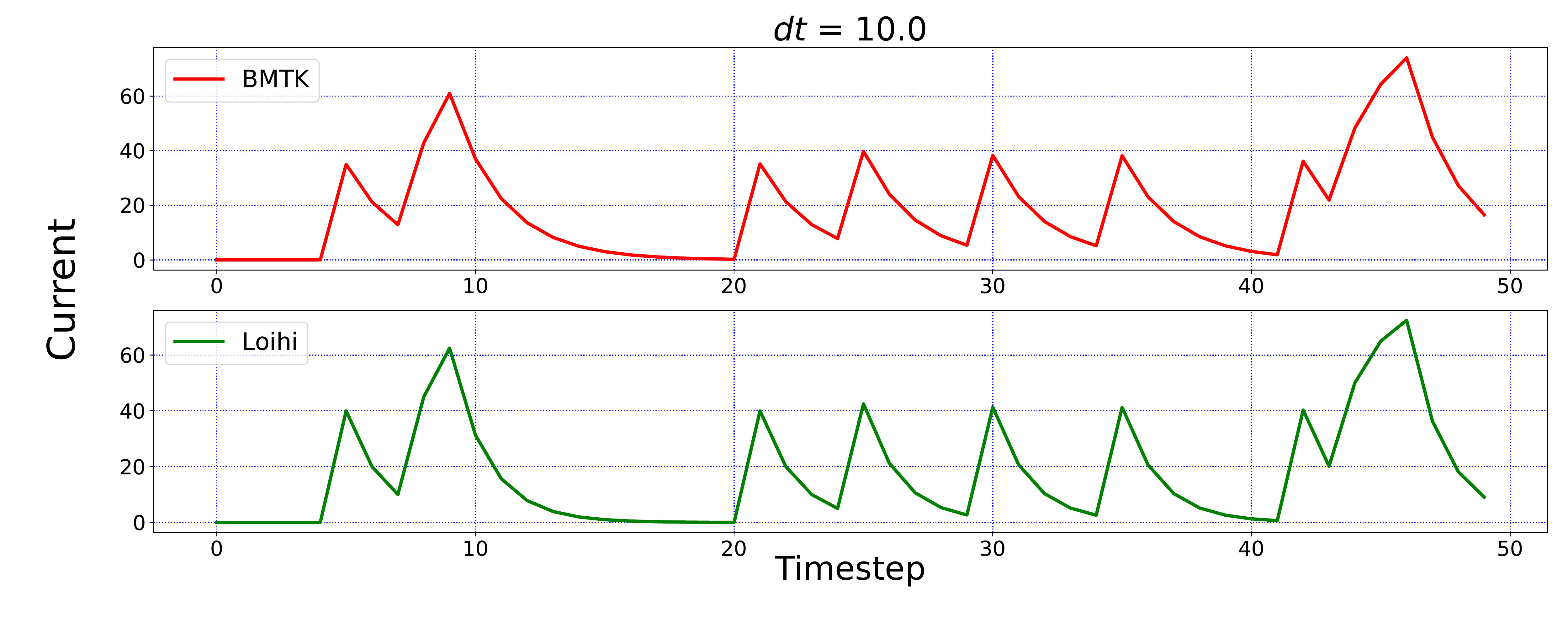}
    \end{minipage}
   \caption{Comparison of membrane potential and current plots with different temporal precisions in Loihi - (a) $dt = 0.1$ (b) $dt = 1.0$ (c) $dt = 10.0$. For $dt=10.0$, number of time-steps are 50 and for $dt=0.1$, number of time-steps are 5000.}
\label{fig:temp_prec_volt_curr}
\end{figure}

We check the results for $dt=10.0, 1.0  \text{ and } 0.1 (ms/timestep).$ As mentioned earlier, we run the simulation for 500$\hspace{0.08cm}ms$, thus the corresponding number of time steps in Loihi for $dt=10$ and $dt=0.1$ becomes 50 and 5000 respectively, and for $dt=1.0$ it remains the same. Figure \ref{fig:temp_prec_volt_curr} illustrates the related Loihi simulation for membrane potential and current.

\noindent \textbf{{\textit{Error comparison for temporal precision}}}

\begin{figure}[htb]
   \begin{minipage}[t]{0.36\textwidth}
        \centering
        \includegraphics[width=\textwidth]{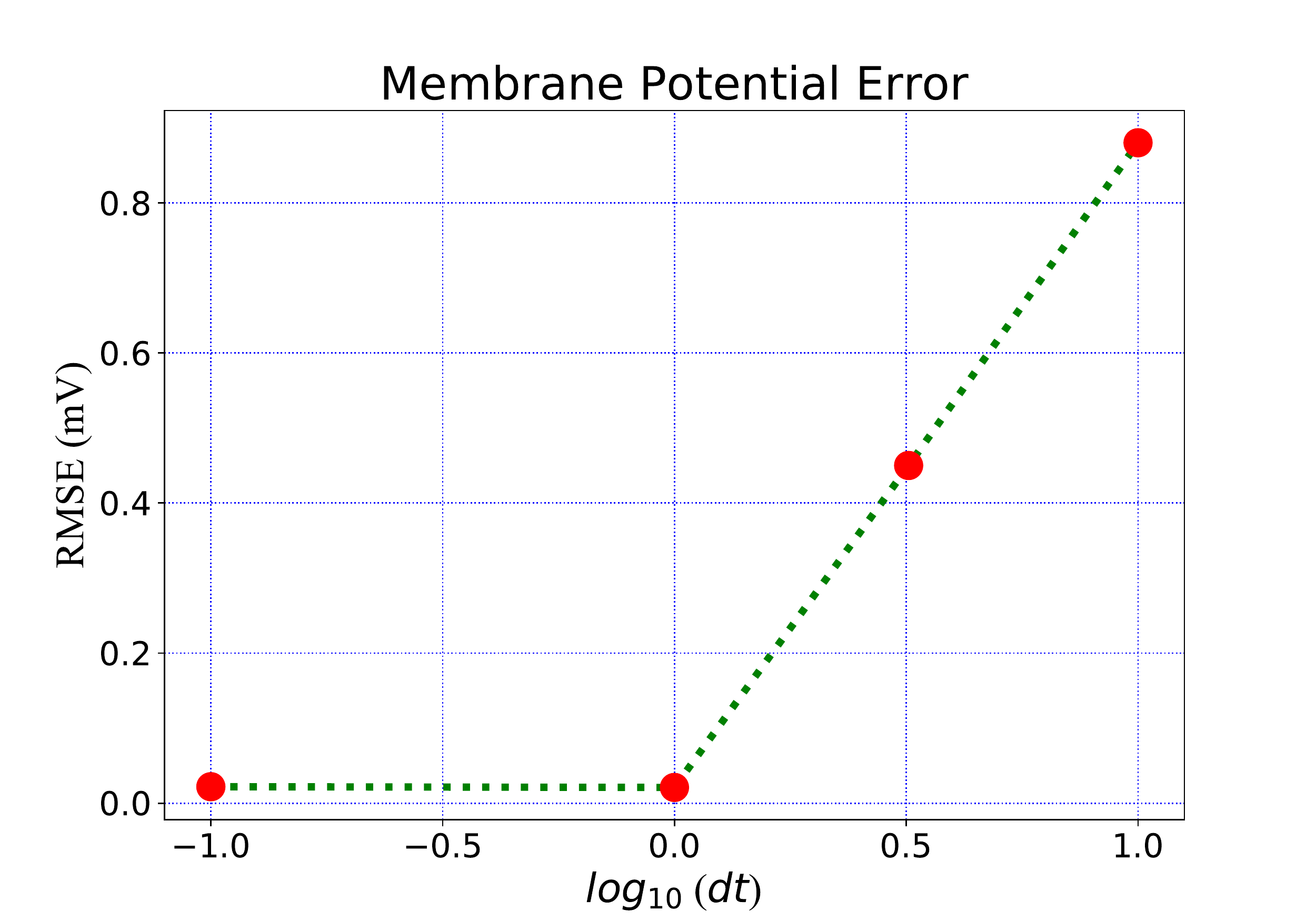}
    \end{minipage}  
    \hspace{1cm}
    \begin{minipage}[t]{0.36\textwidth}
        \centering
        \includegraphics[width=\textwidth]{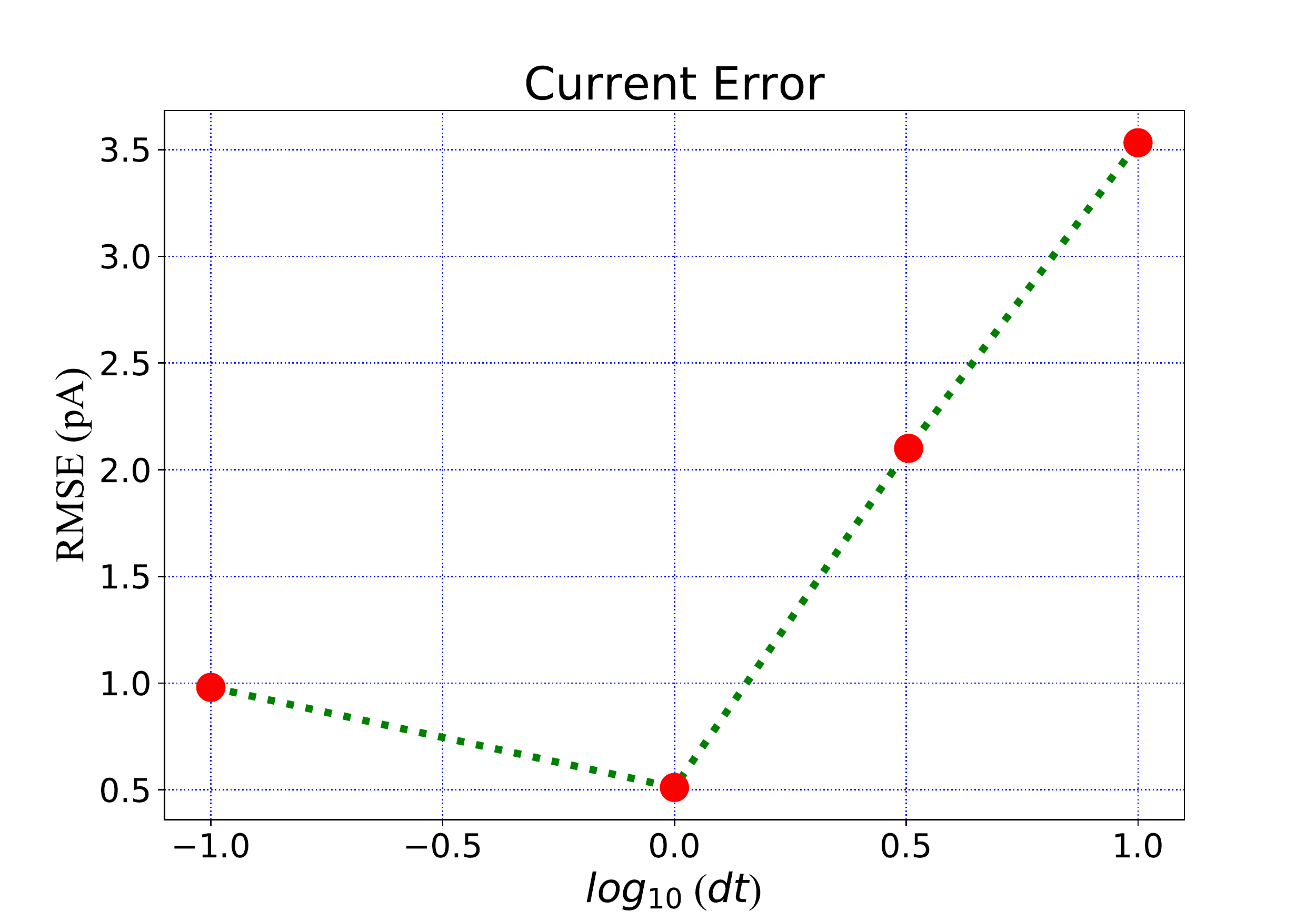}
    \end{minipage}
    \caption{Error comparison for different temporal precisions. In both panels, the RMSE for the corresponding state is plotted against the $log$ of the temporal precision $dt$.}
    \label{fig:temp_error}
\end{figure}

We compare the simulations in Loihi with different temporal precisions against the simulations in BMTK. We calculate the RMSE to be able to deduce the result. As can be seen from Figure \ref{fig:temp_error}, the error is lowest when 1$\hspace{0.08cm}ms$ of simulation time in BMTK equates to 1 time-step in Loihi for membrane potential and current. Thus, for the LIF model simulations, Loihi hardware time is close to the actual timescale of the simulations. 

\subsubsection{Simulation Using Varied Voltage Precision}

\begin{figure}[htb]
   \begin{minipage}[t]{0.48\textwidth}
        \centering
        \includegraphics[width=\textwidth]{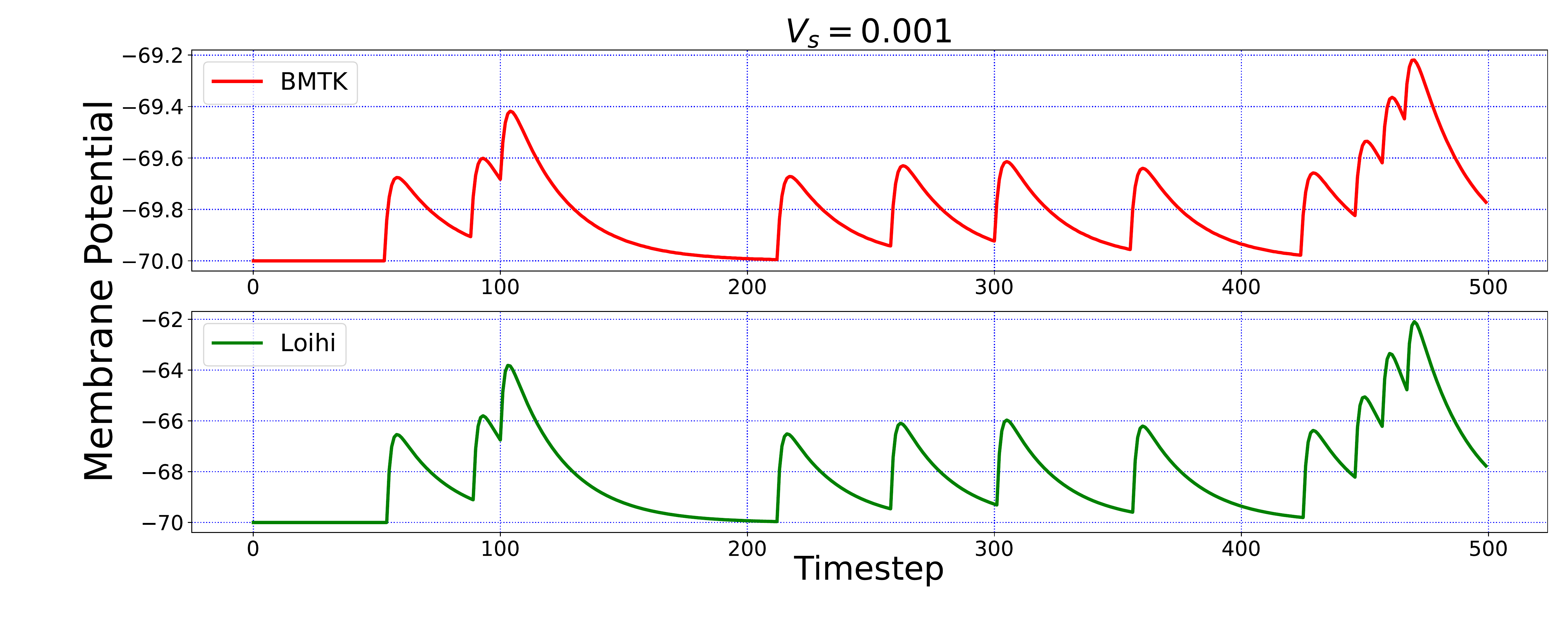}
    \end{minipage}  
    \hfill
    \begin{minipage}[t]{0.48\textwidth}
        \centering
        \includegraphics[width=\textwidth]{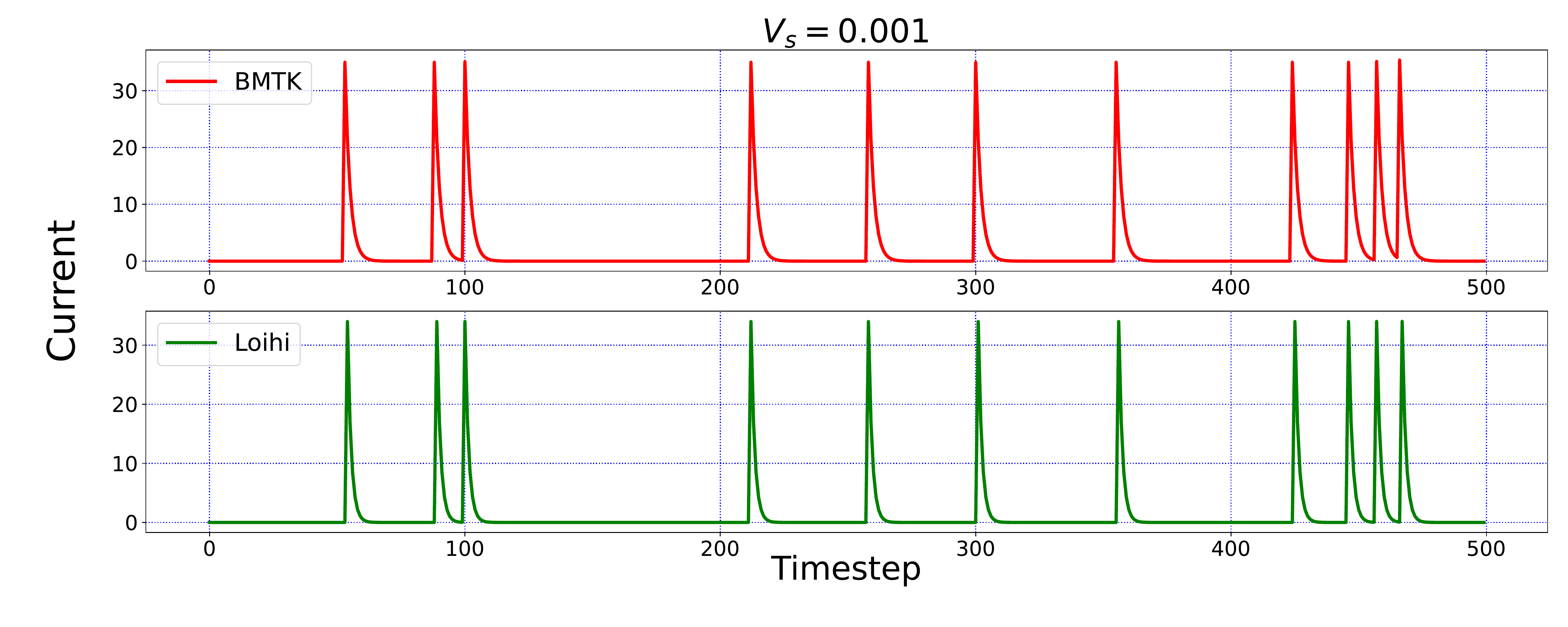}
    \end{minipage}
    \newline
    \begin{minipage}[t]{0.48\textwidth}
        \centering
        \includegraphics[width=\textwidth]{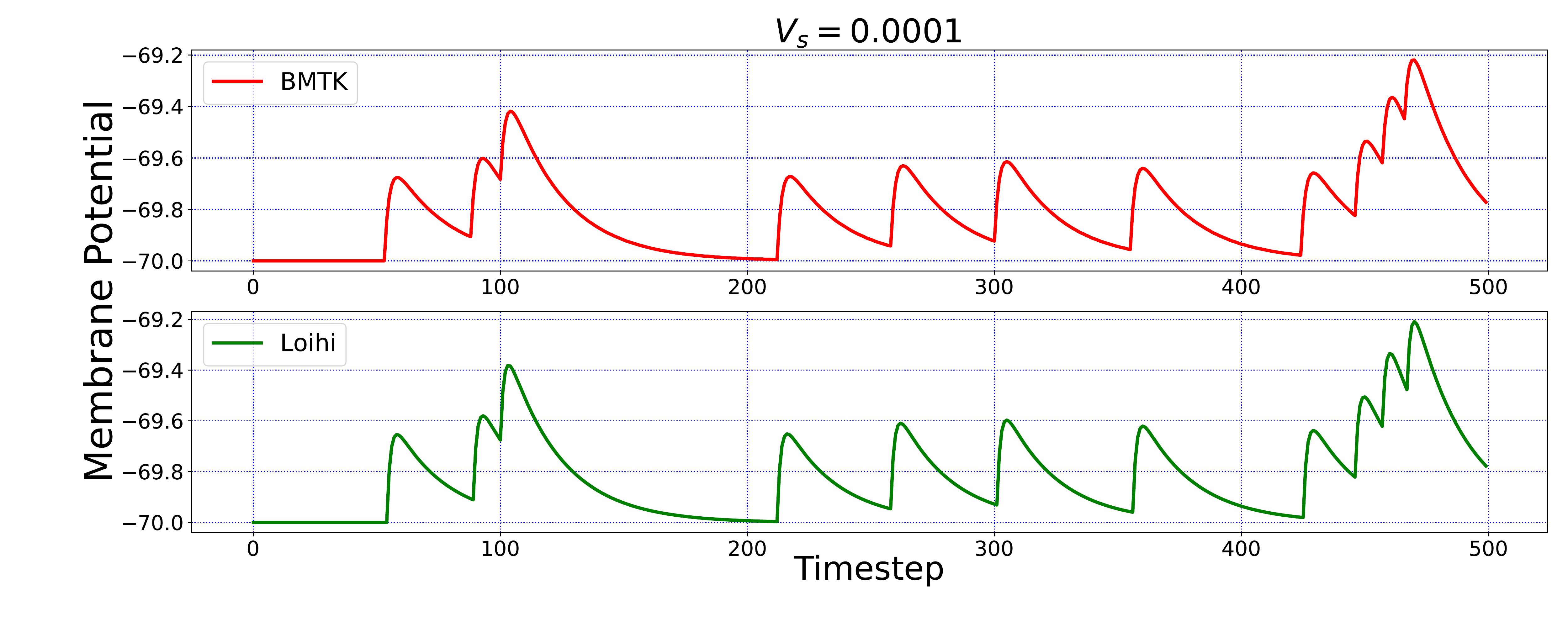}
    \end{minipage}  
    \hfill
    \begin{minipage}[t]{0.48\textwidth}
        \centering
        \includegraphics[width=\textwidth]{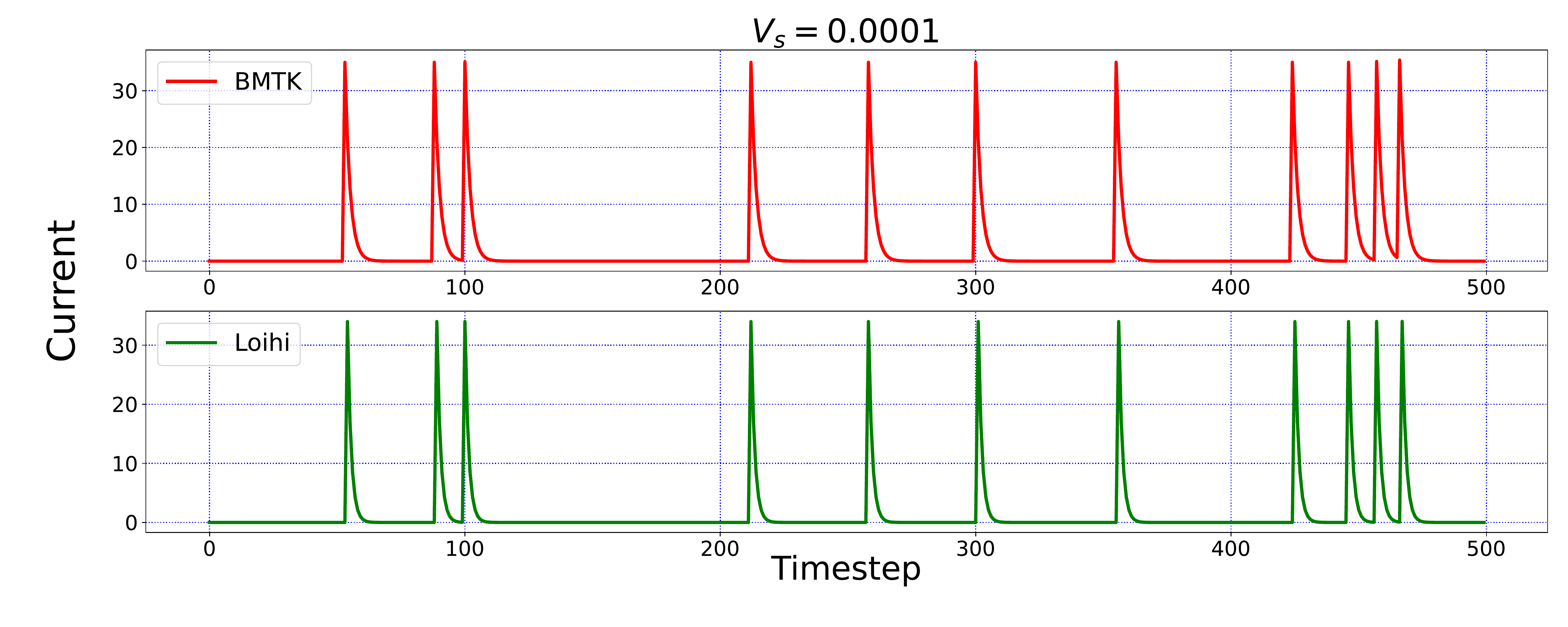}
    \end{minipage}
    \newline
    \begin{minipage}[t]{0.48\textwidth}
        \centering
        \includegraphics[width=\textwidth]{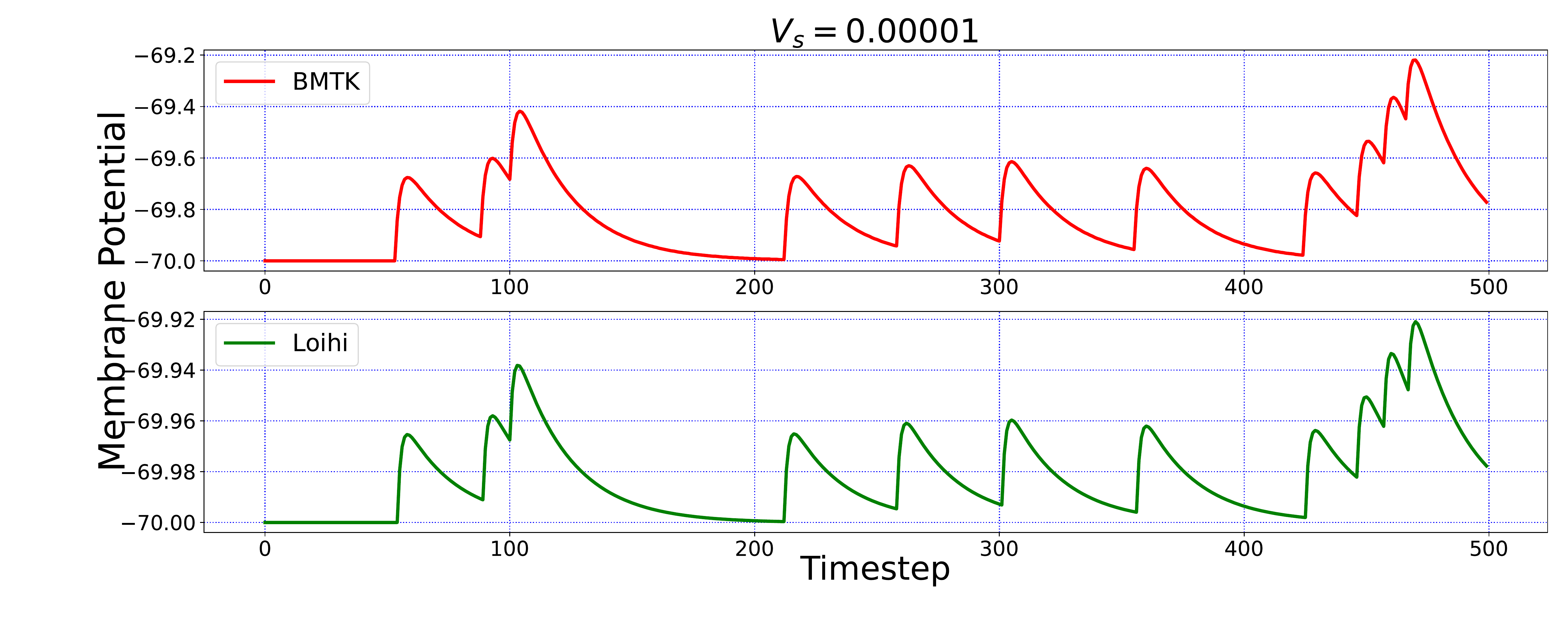}
    \end{minipage}  
    \hfill
    \begin{minipage}[t]{0.48\textwidth}
        \centering
        \includegraphics[width=\textwidth]{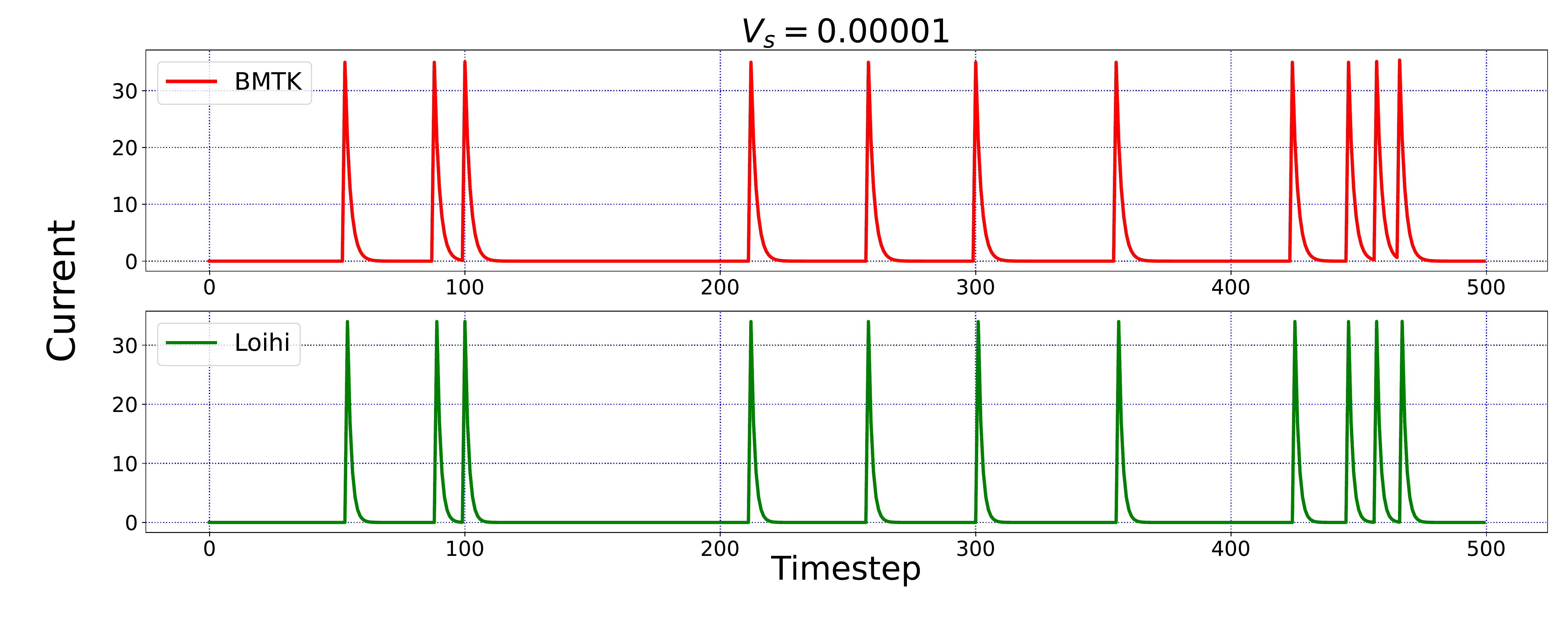}
    \end{minipage}
   \caption{Comparison of membrane potential and current plots with different voltage precisions - \newline (a) $V_{s} = 1.0 \times 10^{-3}$ (b) $V_{s} = 1.0 \times 10^{-4}$ (c) $V_{s} = 1.0 \times  10^{-5}.$}
\label{fig:vol_prec_volt_curr}
\end{figure}

We repeat the precision study by changing the voltage precision values using the re-scaling parameter $V_{s}$. To check different precision results, we try $1$K$/$mV, $10$K$/$mV \text{ and } $100$K$/$mV (state level/mV) by using $V_{s} = 1.0 \times 10^{-3}, 1.0 \times 10^{-4}$ and $1.0 \times 10^{-5}$ respectively. Figure \ref{fig:vol_prec_volt_curr} illustrates the neuron state simulations based on different voltage precisions.

\noindent \textbf{\textit{Error comparison for voltage precision}}

\begin{figure}[htb]
   \begin{minipage}[t]{0.36\textwidth}
        \centering
        \includegraphics[width=\textwidth]{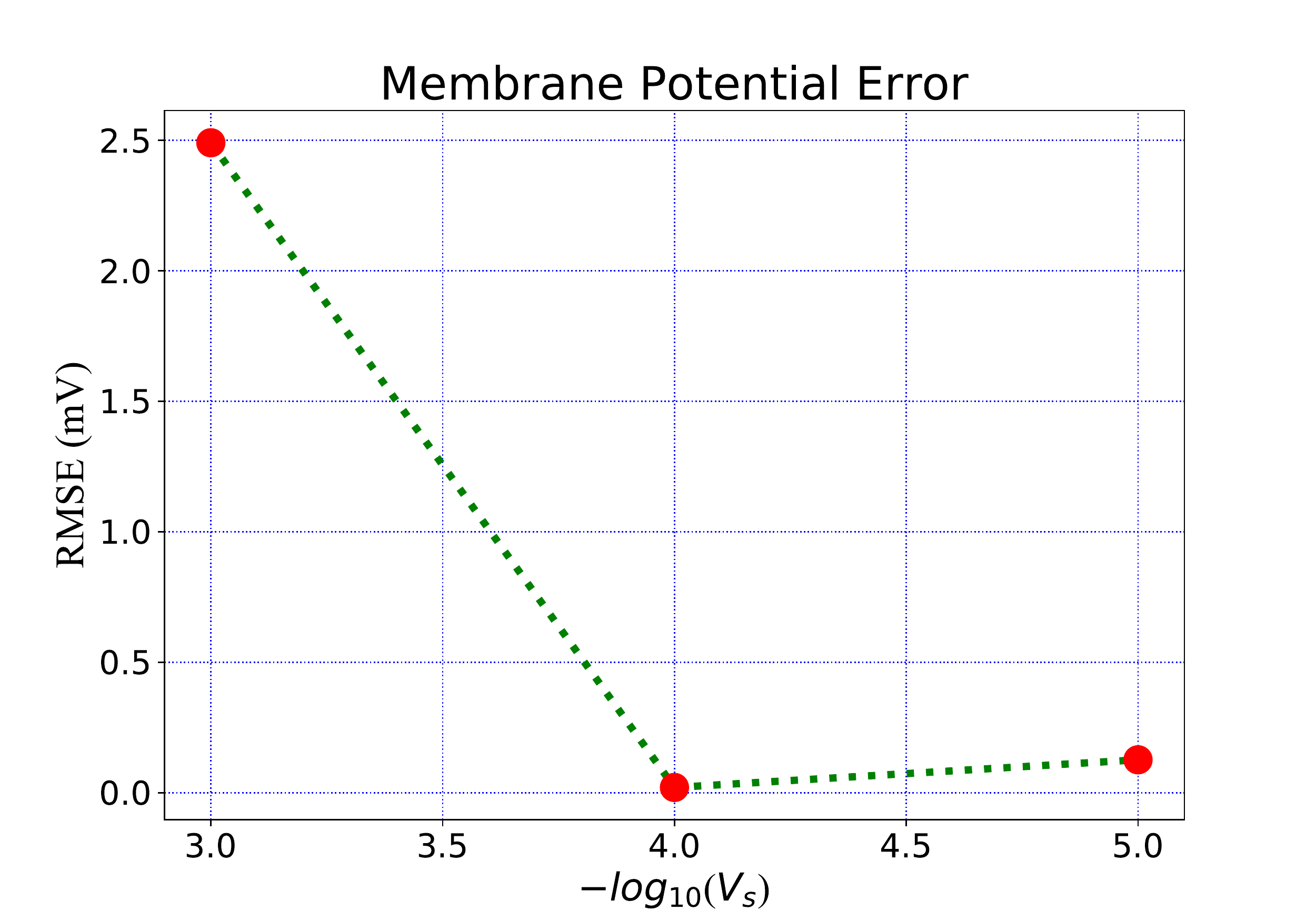}
    \end{minipage}  
    \hspace{1cm}
    \begin{minipage}[t]{0.36\textwidth}
        \centering
        \includegraphics[width=\textwidth]{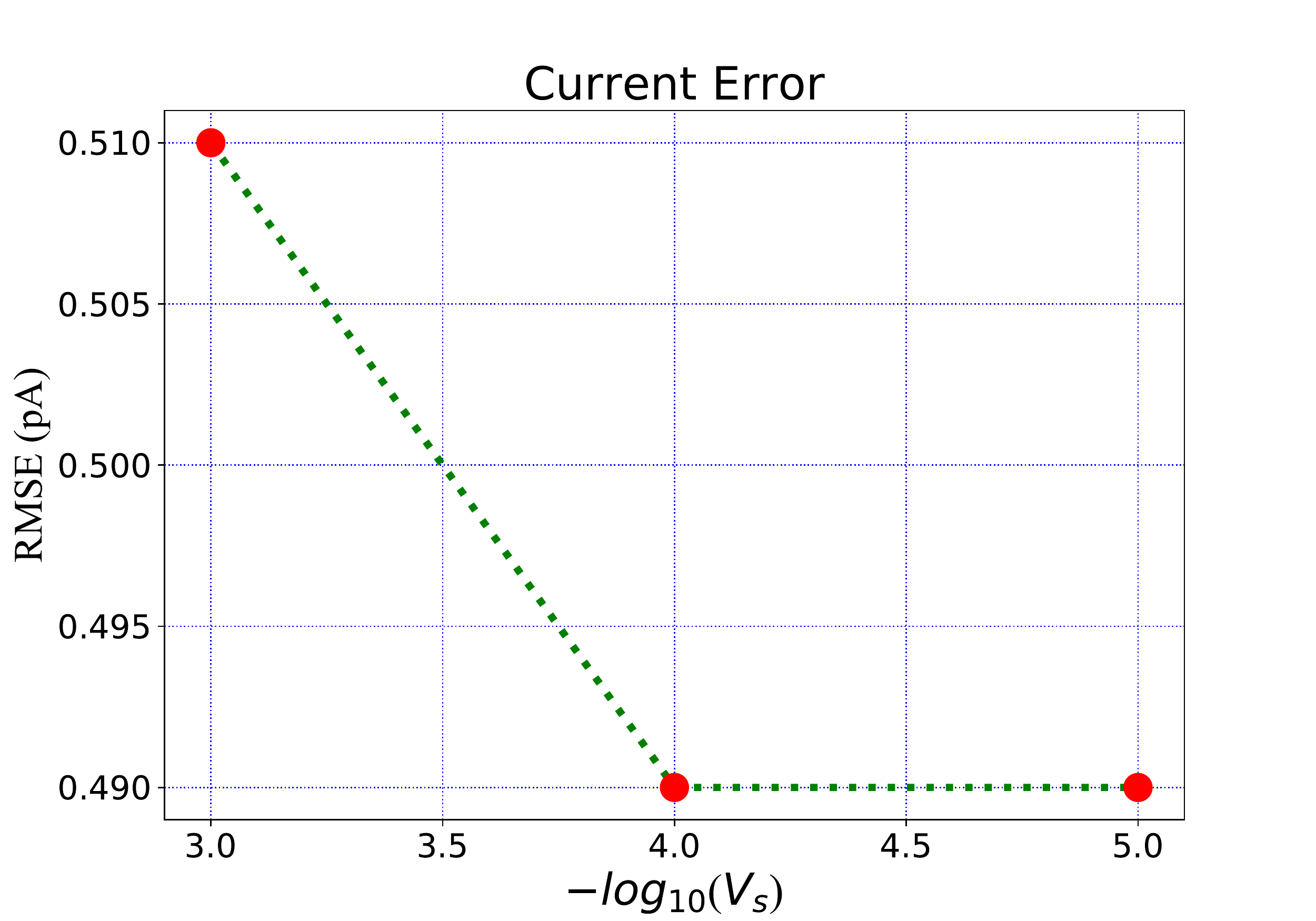}
    \end{minipage}
    \caption{Error comparison for different membrane potential precisions. In both panels, the RMSE for the corresponding state is plotted against the $-log$ of the voltage scale $V_s$.}
    \label{fig:volt_error}
\end{figure}

 As can be seen from Figure \ref{fig:volt_error}, for membrane potential - error decreases significantly as the precision increases from $V_{s}=1.0 \times 10^{-3}$ to $V_{s}=1.0 \times 10^{-4}$. However, the error is extremely small for the current simulation and remains the same for $V_{s}=1.0 \times 10^{-4}$ and $V_{s}=1.0 \times {10}^{-5}$.

In conclusion, the effect of precision on both scales depends on the model parameters and the information needing to be preserved. However, there are important performance differences. Increased voltage precision is essentially free, as it does not tax the hardware resources any further, and the sole risk is from computation overflow in cases of the Loihi voltage state nearing the precision of the voltage register. Increased time precision on the other hand has two important drawbacks: it increases simulation time (proportionately to increased precision), and it decreases the range of voltage decay timescale that can be represented (again, proportionately to increased precision). Thus, the choice of simulation time step and corresponding precision should be weighed against these trade offs.

\subsection{Simulation of Larger Networks}
After establishing and verifying the calibrated Loihi parameters, we extend our network to larger number of neurons. Currently, we have successfully implemented a network of $\sim$10,000 neurons. It is done by increasing the number of cores that are involved in the processing of the network, as each core supports only 1024 neurons. 

\begin{figure}[htb]
   \begin{minipage}[t]{\textwidth}
        \centering
        \includegraphics[width=0.6\textwidth]{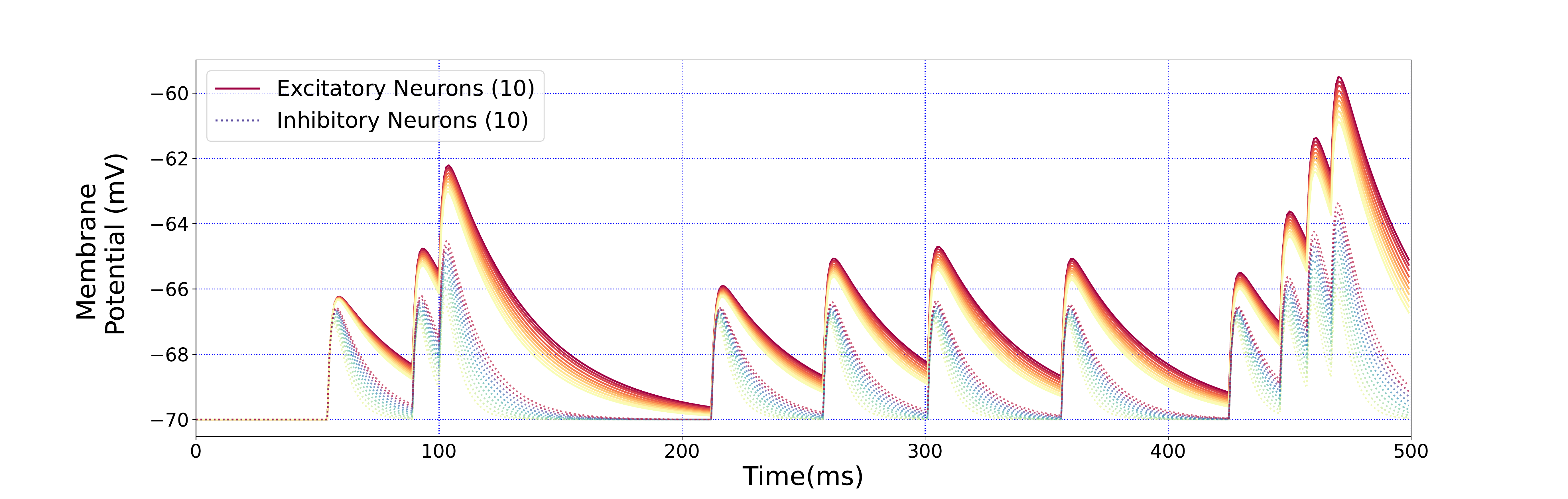}
        \subcaption{BMTK network of 20 neurons}
    \end{minipage}  
    \newline
    \begin{minipage}[t]{\textwidth}
        \centering
        \includegraphics[width=0.6\textwidth]{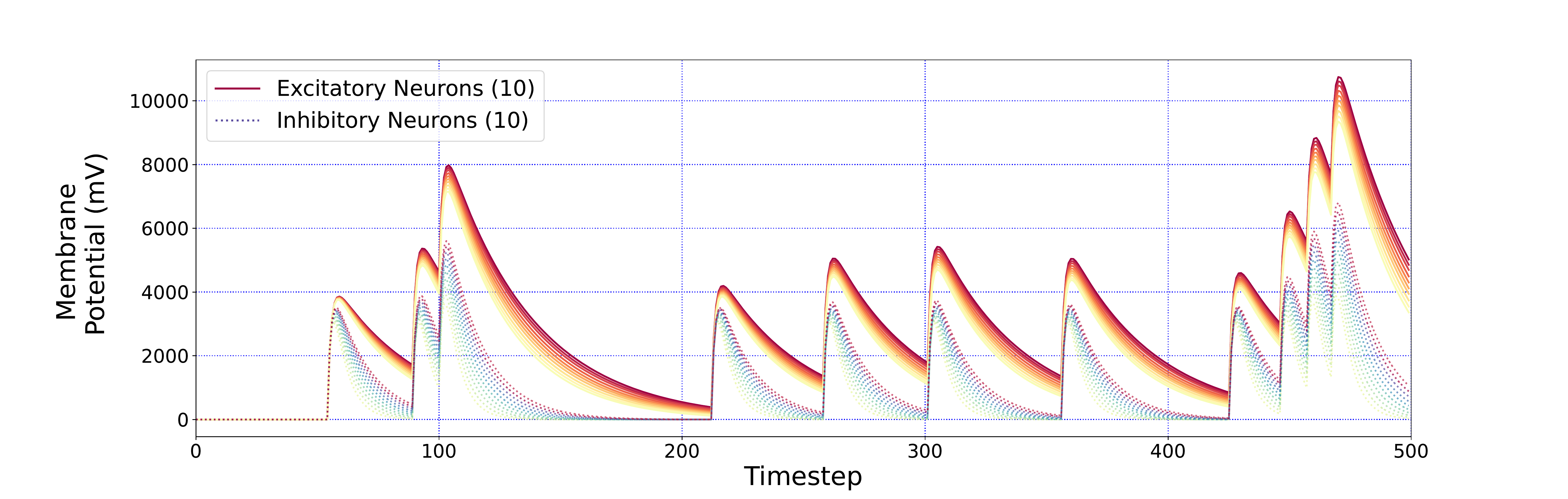}
        \subcaption{Loihi network of 20 neurons}
    \end{minipage}
    \caption{Loihi replicates large network response of BMTK}
    \label{fig:large_network}
\end{figure}

Figure \ref{fig:large_network} illustrates an equivalent simulation for a network of 20 neurons between BMTK and Loihi indicating  that Loihi is capable of emulating larger BMTK networks. (Figure \ref{fig:large_network} (b) shows the raw membrane potential values for Loihi, before the inverse mapping in equation (\ref{eq:inverse}) is applied). Here, we have a correlation of 0.99985 with an RMSE of 0.24 $\times 10^{-3}$ mV/ $ms$. This also validates the fact that the  calibration of parameters for smaller networks done earlier is valid. This lays the foundation for building bigger and more complex networks on Loihi.

\subsection{Comparison of Performance}
The performance of Loihi far exceeds that of BMTK. For smaller networks, the time taken to build, compile and execute the networks is approximately similar in both the platforms. However, as the networks get larger, BMTK takes much longer than Loihi to  complete the simulations.

\begin{figure}[htb]
   \begin{minipage}[t]{0.46\textwidth}
        \centering
        \includegraphics[width=\textwidth]{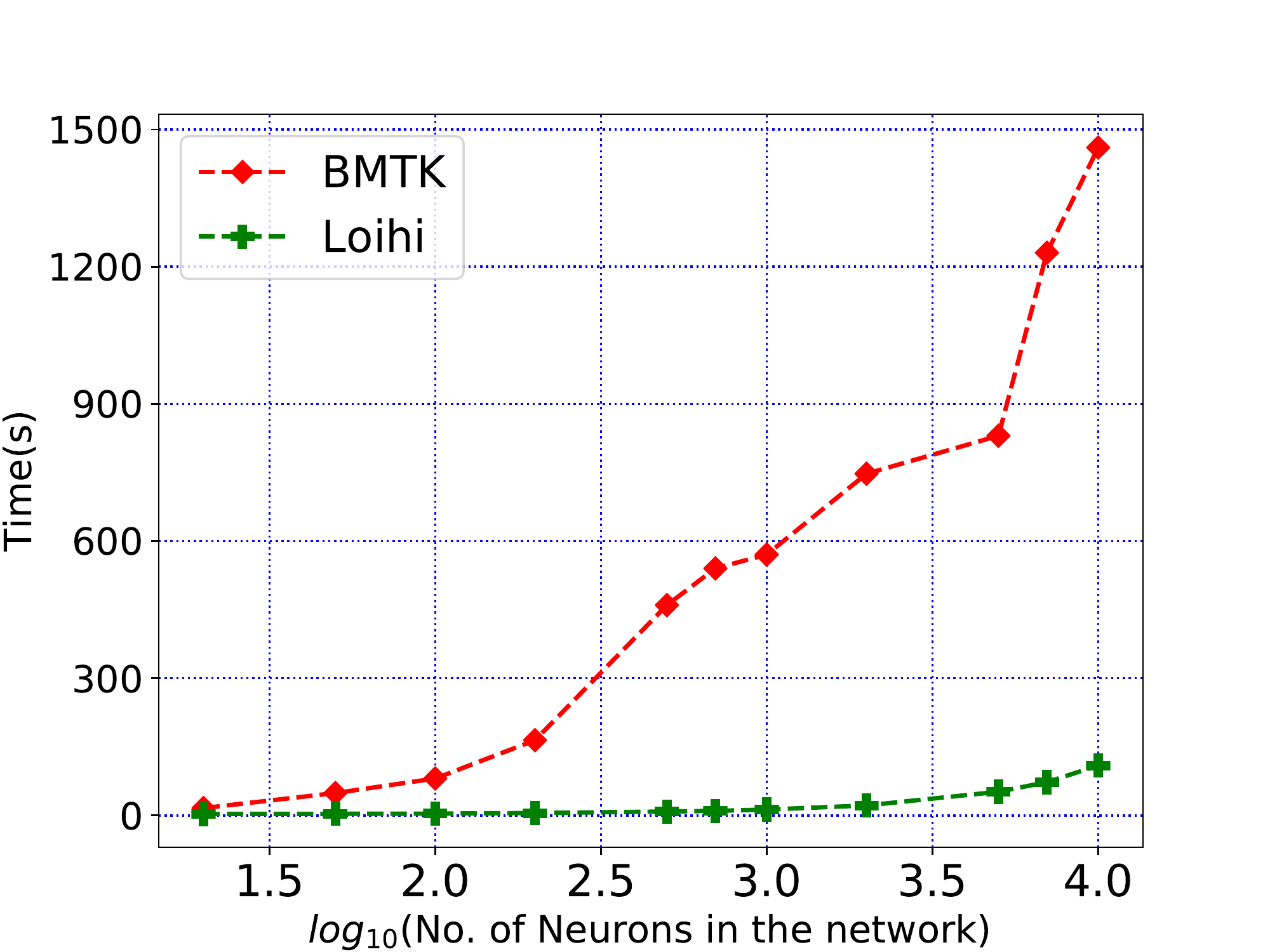}
        \subcaption{Simulation time vs. logarithm of network size}
    \end{minipage}  
    \hspace{0.5cm}
    \begin{minipage}[t]{0.46\textwidth}
        \centering
        \includegraphics[width=\textwidth]{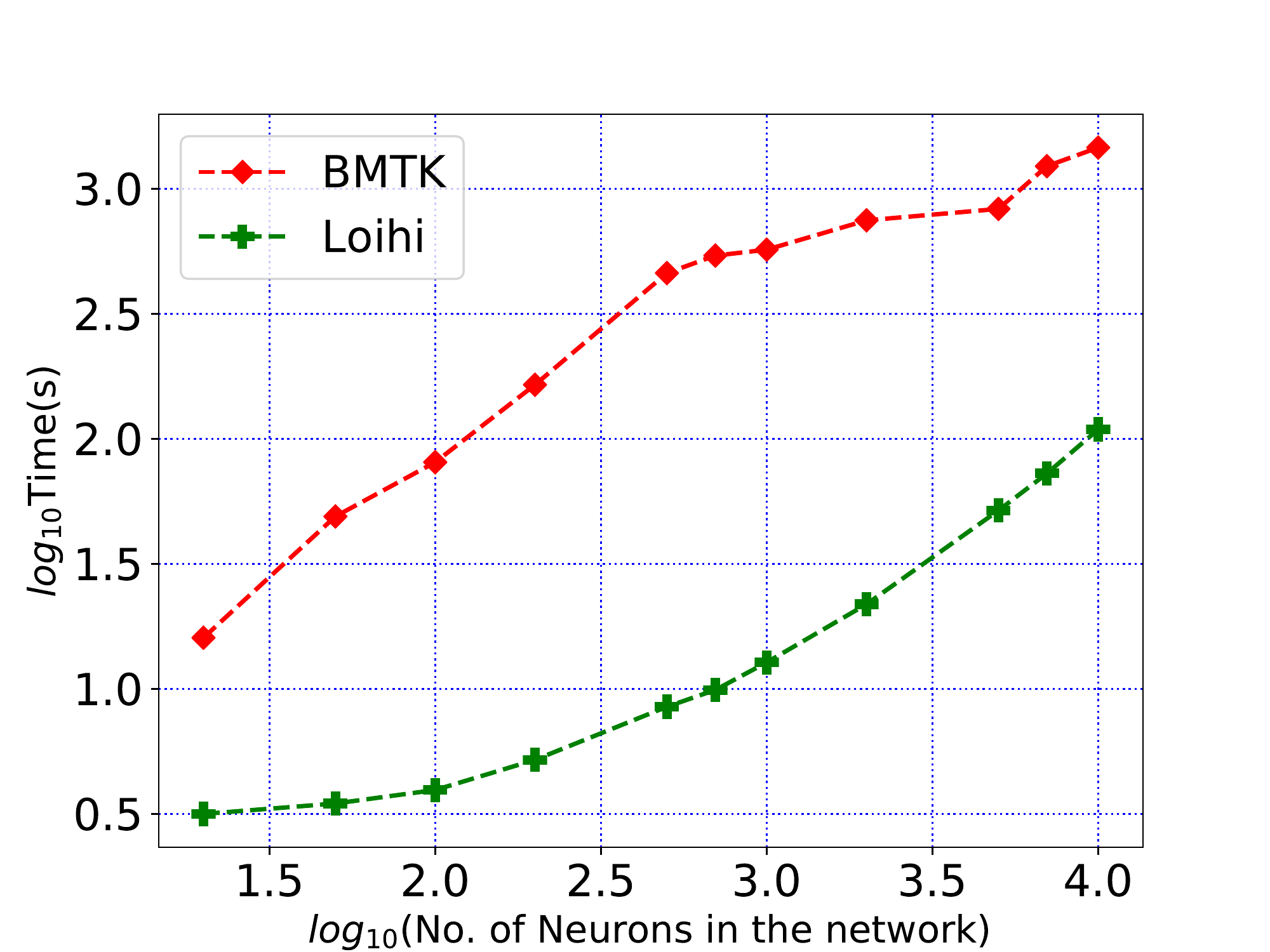}
        \subcaption{Logarithm of simulation time vs. logarithm of network size}
    \end{minipage}
    \caption{Performance comparison between BMTK and Loihi for network sizes ranging from 1 to 10000, expressed in $log$ scale.}
    \label{fig:time}
\end{figure}

Figure \ref{fig:time} compares the time taken by BMTK and Loihi to complete the simulations. We measure the system and user CPU time of each run, excluding the time elapsed in queue or sleep. 

The irregularities in the BMTK curve can be attributed to the initial loading of the data through `h5py' files which have an initial large load time but scale well as the network gets larger. Moreover, NEST initializes neurons in batches and memory pre-allocation follows the $log_{2}$ scale. Care was taken to run the simulations on a single core on the same node to ensure no variability in terms of system performance.

Evidently, Loihi could prove to be a huge advantage as we aim to undertake the mammoth task of emulating the mouse primary visual cortex.

\section{Conclusion}
Inspired by the brain, neuromorphic computing holds great potential in tackling tasks with extremely low power and high efficiency. Many large-scale efforts including the TrueNorth, SpiNNaker and BrainScaleS have been demonstrated as a tool for neural simulations, each replete with its own strengths and constraints. Fabricated on Intel’s 14nm technology node, Loihi is a forward-looking and continuously evolving state-of-the-art architecture for modeling spiking neural networks in silicon. As opposed to its predecessors, Loihi encompasses a wide range of novel features such as hierarchical connectivity, dendritic compartments, synaptic delays and programming synaptic learning rule. These features, together with solid SDK support by Intel, and a growing research community, make Loihi an effective platform to explore a wealth of neuromorphic features in more detail than before.

In this work, we have demonstrated that Loihi is capable of replicating the continuous dynamics of point neuronal models with high degree of precision and does so with much greater efficiency in terms of time and energy. The work comes with its challenges as simulations built on the conventional chips cannot be trivially mapped to the neuromorphic platform as its architecture differs remarkably from the conventional hardware. 
Classical simulations from the Brain Modeling Toolkit (BMTK) developed by the Allen Institute of Brain Science (AIBS) serves as the foundation of our neuromorphic validation.

 For comparison between the conventional and the neuromorphic platforms, we use both qualitative and quantitative measures. It can be seen that Loihi replicates BMTK very closely in terms of both membrane potential and current, the two state variables on which the Loihi LIF model evolves. We use different validation methods and quantitative measures to assess the equivalence and identify sets of parameters which maximize precision while retaining high performance levels. 
Furthermore, simulation results indicate Loihi is highly efficient in terms of speed and scalability as compared to BMTK. As indicated by the Figure \ref{fig:time}, Loihi is almost a thousand times faster than BMTK for smaller networks and scales markedly well with network size.

In conclusion, this work demonstrates that classical simulations based on Generalized Leaky Integrate-and-Fire (GLIF) point neuronal models can be successfully replicated on Loihi with an exceptional degree of precision. Additionally, the high efficiency and low-power consumption of the neuromorphic platform with increasing network size paves the way for a complete replication of the mouse visual cortex dynamics, comprising hundreds of thousands of neurons and millions of synapses.

\section*{Acknowledgements}
We thank Dr. Kael Dai for his helpful suggestions and advice regarding the use of BMTK. We thank Intel for providing us  with access to Loihi and the Intel technical support team for helpful feedback on technical issues.

\newpage
\bibliography{ref.bib}
\end{document}